\documentclass[sigconf]{acmart}
\usepackage{caption}
\usepackage{subcaption}
\usepackage{multirow}
\AtBeginDocument{%
  \providecommand\BibTeX{{%
    \normalfont B\kern-0.5em{\scshape i\kern-0.25em b}\kern-0.8em\TeX}}}

\setcopyright{acmcopyright}
\acmYear{2022}

\acmConference[Conference acronym 'XX]{xxx}{xxx}{xxx }
\acmBooktitle{xxx,
  xxx, xxx} 

\begin{document}

\title{How News Evolves? Modeling News Text and Coverage using Graphs and Hawkes Process}


\author{Honggen Zhang}
\affiliation{%
  \institution{University of Hawai'i at M\=anoa}
  \city{Honolulu}
  \country{USA}}
\email{honggen@hawaii.edu}

\author{June Zhang}
\affiliation{%
  \institution{University of Hawai'i at M\=anoa}
  \city{Honolulu}
  \country{USA}}
\email{zjz@hawaii.edu}


\begin{abstract}
Monitoring news content automatically is an important problem. News content, unlike traditional text, has a temporal component. However, few works have explored the combination of natural language processing and dynamic system models. One reason is that it is challenging to mathematically model the nuances of natural language. In this paper, we discuss how we built a novel dataset of news articles collected over time. Then, we present a method of converting news text collected over time to a sequence of directed multi-graphs, which represent semantic triples ($\text{Subject} \rightarrow \text{Predicate} \rightarrow \text{Object}$). We model the dynamics of specific topological changes in these graphs using a set of multivariate count series, which we fit the discrete-time Hawkes process. With our real-world data, we show that the multivariate time series contain both dynamic information of how many articles/words were published each day and semantic information of the content of the articles. This yields novel insights into how news events are covered. We show with the experiment that our approach can be used to infer from a sequence of news articles if the articles were published by major or entertainment news outlets.
\end{abstract}




\maketitle

\section{Introduction}\label{introduction }
 The rise of internet news media generates a lot of news streams. Unlike general text, news text is inherently temporal, comprising of a series of events over time and space. Several problems are associated with the automatic analysis of news text such as event detection, relation detection, and fake news detection. Researchers working in Temporal Text Mining use tools to automatically extract events and to cluster and organize them by temporal relationships. However, these works do not model how the events change over time with a mathematical model. Dynamic information has proven to be useful in natural language processing (NLP) tasks on events that evolve with time. Classifiers for detecting fake tweets use a combination of text features and count information of tweets and retweets over time. 

In this paper, we present a framework of using NLP and time series analysis to model how news coverage of a particular event changes over time. The time series we generate from the text contains both dynamic information such as how many articles were published on the event and semantic information of what the articles discussed.

First, we collected news articles from the internet related to specific events using Event Registry. Two events are discussed in this paper: 1) the \textbf{Alleged assault of Jussie Smollett} and 2) \textbf{Ukraine International Airlines Flight 752 crashed in Iran}. The dataset, collected from 3,145 different news sources, contains 44,403 news articles with the complete raw text, publication time, and metadata information. The dataset is available on Github\footnote{https://github.com/honggen-zhang/News-Evolve-on-DHP}.
 
 We process the article text with ReVerb \cite{fader2011identifying} to extract semantic triples with the pattern $\text{Subject} \rightarrow \text{Predicate} \rightarrow \text{Object}$. We represent the triples as the nodes and edges of a Resource Description Framework (RDF) graph (i.e., directed multi-graph with words and phrases associated with both the nodes and the edges). A stream of news articles can be converted to a sequence of RDF graphs.
 
 The dynamics of how the RDF graphs change over time contain semantic (i.e., what happened), linguistic (i.e., how events were described/interpreted), and coverage (i.e., how many articles were published) information. Because it is difficult to mathematically model dynamic graphs, we keep track of the count of specific topological changes to the triples in the initial RDF graph instead. We modeled this induced set of multivariate count time series using a discrete-time Hawkes process.

Section~\ref{sec:related} will review related prior work. We discuss our data collection and cleaning methods in Section~\ref{sec:data}. Section~\ref{ProDef} details how we transformed text data into a sequence of RDF graphs and to a collection multivariate count time series induced by the semantic triples. We introduce the discrete-time Hawkes process in Section~\ref{sec:DHP}. We discuss our experiments in fitting the collected data to the Hawkes process and the analysis of the learned parameters in Section~\ref{sec:exp}. We show that our approach shows a different dynamic between articles published by major news outlets (e.g., Politico, BBC, New York Times, NPR, Reuters) and by entertainment news outlets (e.g., TMZ, People, E!Online, Global News, The Sun, Pop Sugar). The learned parameters of the Hawkes process can be used to distinguish if a set of articles published by unknown outlets came from major or entertainment news outlets.

\section{Related work}\label{sec:related}
\subsection{Text mining}
Works in Temporal Text Mining (TTM) are primarily focused on either event detection or relation detection. Works in event detection focus on automatically finding events from sets of news articles \cite{allan1998topic, allan1998line}. In some work, the events are then placed in some sort of temporal order and visualized as a directed graph. In \cite{ghalandari2020examining}, the news articles related to specific events are also automatically summarized. Various clustering methods have been used to gather multiple articles related to a specific event together \cite{brants2003system, zhao2007temporal}.

Work such as the Event Threading model in \cite{nallapati2004event} and the Story Forest in \cite{liu2020story} place more emphasis on discovering the relationships (usually temporal) that connect multiple events together. The temporal relationships are visualized as a directed graph and are used to track the dynamic of events. Works that have used graphs to describe the evolutionary pattern of specific events are \cite{mei2005discovering, das2011dynamic, yang2009discovering, spitz2019topexnet}. Usually, evolution is characterized by some sort of distance measurement between pairs of events.

\subsection{Hawkes process}
The Hawkes process is a point process whose realization consists of discrete events localized in time \cite{hawkes1971spectra1}. It is a self-exciting process in that the rate of occurrence depends on the history of the process. For example, a history of a high number of occurrences in a short period of time will increase the probability of the event occurring in the near future. Hawkes processes are often used to model time series data with `bursty' dynamics.

Recently, several works have used the Hawkes process to build classifiers to detect fake tweets \cite{ kobayashi2016tideh, farajtabar2017coevolve}. In Dutta et al.'s HawkesEye classifier \cite{dutta2020hawkeseye}, dynamics parameters of the learned Hawkes process are combined with semantic information from the text to generate features to train the classifiers. Reference~\cite{kobayashi2016tideh} used the Hawkes process to build a normative model of the dynamics of tweets, which can then be used for anomaly detection.

\section{Data Collection and Extraction}\label{sec:data}
\begin{figure}[htb]
    \centering
    \includegraphics[width=1.0\columnwidth]{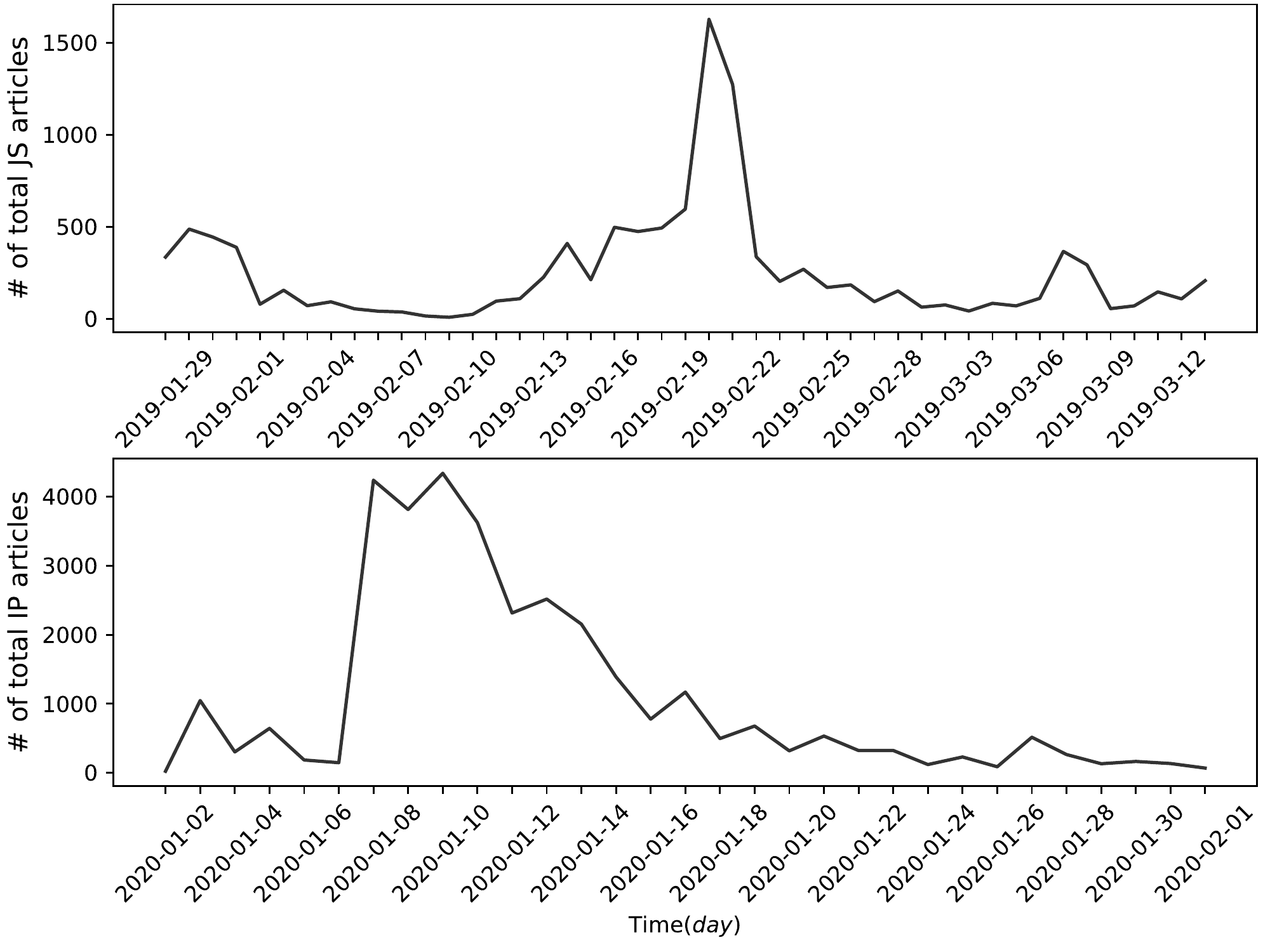}
    \caption{Number of Articles Collected per Day of Dataset JS and Dataset IP}
    \label{article of total}
\end{figure}

We scrapped news articles from the news collection platform Event Registry\footnote{https://eventregistry.org/}. The platform provides a useful Python API, so we can download custom news articles by attributes such as keywords, language, place. In this paper, two events were selected, the \textbf{Alleged assault of Jussie Smollett} and \textbf{Ukraine International Airlines Flight 752 crashed in Iran}. Each news article is scrapped based on specific keywords.
Specifically, we used \textit{Jussie Smollett} and \textit{Iran, Plane} as the keywords to obtain two datasets: JS and IP, respectively. The JS dataset contains 11,340 news articles collected from 1,009 news sources spanning January 28, 2019 to March 14, 2019. The IP dataset contains 33,063 news articles collected from 2,136 news sources spanning January 02, 2020 to February 01, 2020. Figure~\ref{article of total} shows the number of articles collected each day. In addition to the extracted raw text, collected articles also contain (if available): 1) URL of the article, 2) publication date, 3) title, 4) authors, 5) URL of associated image, 6) keywords.

\begin{figure*}[ht]
     \centering
     \hfill
     \begin{subfigure}{0.45\textwidth}
         \centering
         \includegraphics[width=1.0\textwidth]{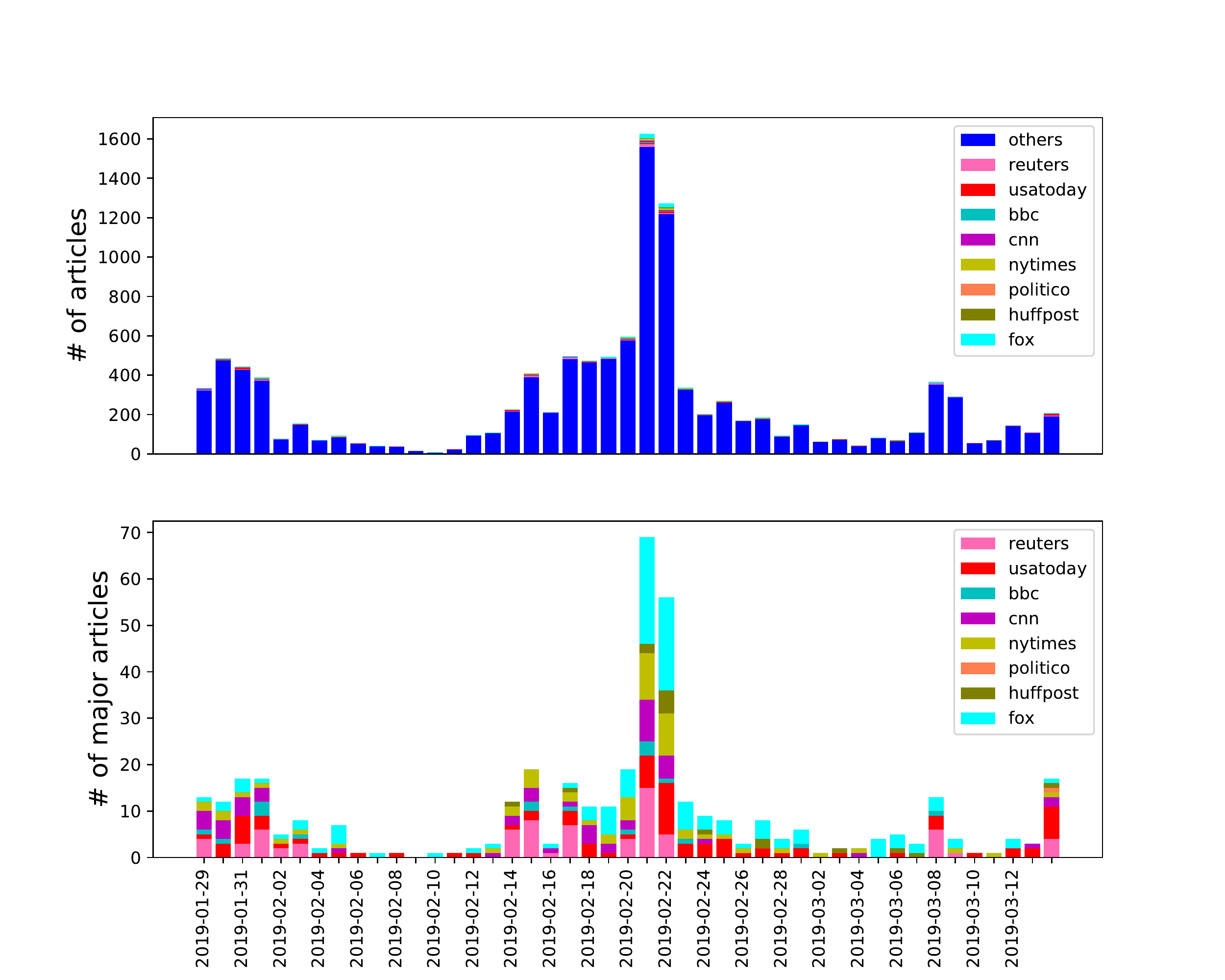}
         \caption{Number of Articles Published by JS }
         \label{JS Bignews}
     \end{subfigure}
     \hfill
     \begin{subfigure}{0.45\textwidth}
         \centering
         \includegraphics[width=1.0\textwidth]{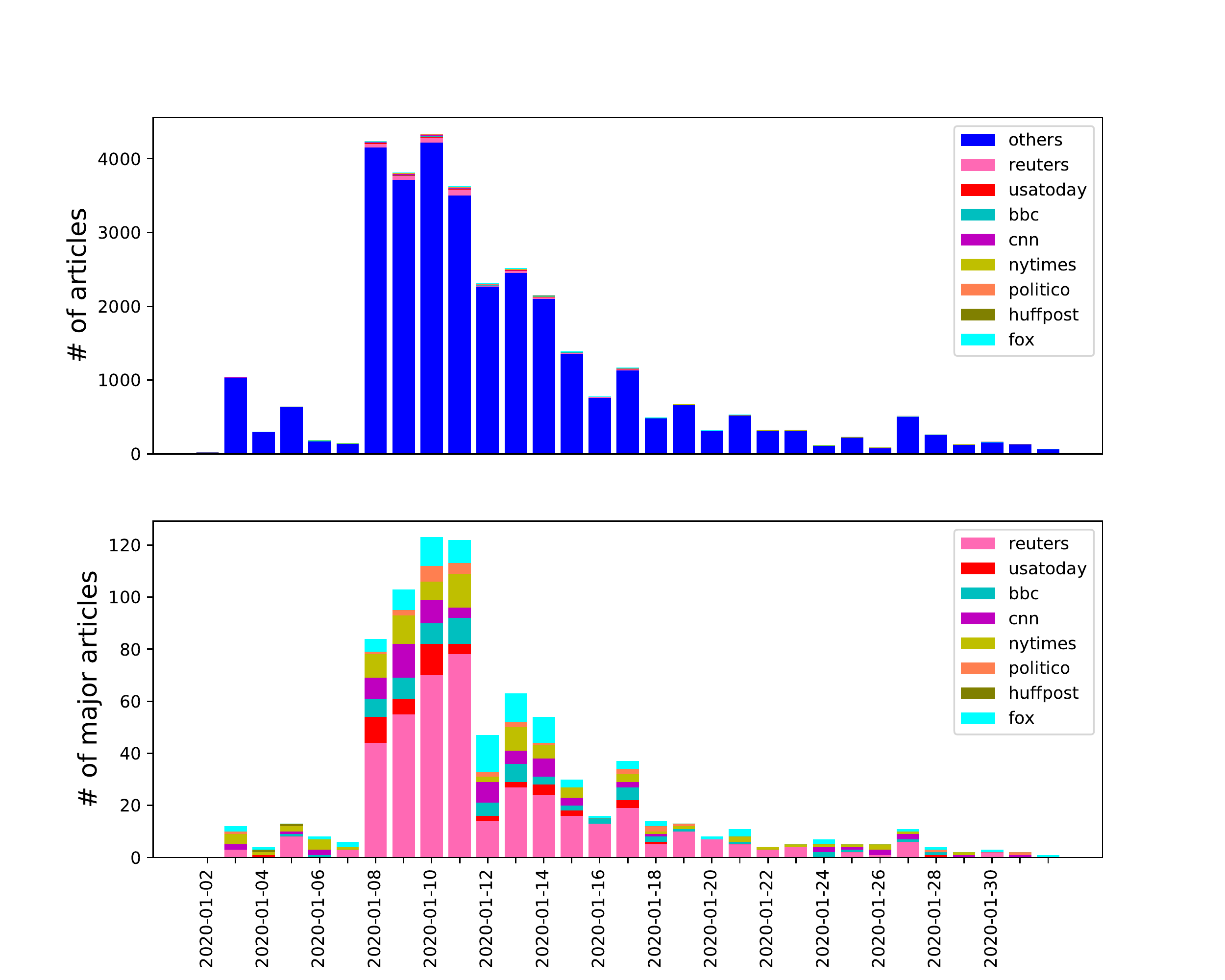}
         \caption{The article count by sources for news IP }
         \label{IP Gossip}
     \end{subfigure}
     \hfill
        \caption{The article count by sources. The bottom figure is the detail of top figure about major news.}
        \label{count of BG}
\end{figure*}

It is also interesting to consider the composition of sources of all these articles pushed onto the internet each day. Only a very small fraction of articles were published from recognizable major news outlets as shown in Figure~\ref{count of BG}.

\subsection{Text to Triples to Graph Representation}

We needed to extract the semantic information from the article text. We used ReVerb \cite{fader2011identifying} to extract semantic triples in the form $\text{Head}(h) \rightarrow \text{Relation}(r) \rightarrow \text{Tail}(t)$. For example, the triple $\text{Barack Obama} \rightarrow \text{born in} \rightarrow \text{Honolulu}$ can be extracted from the sentence \textit{Barack Obama was born in Honolulu}. 

Reverb extracts triples from syntactic constraints based on verb phrases. Therefore, multiple triples may be extracted from a single sentence. Triples follow the basic $\text{Subject} \rightarrow \text{Predicate} \rightarrow \text{Object}$ pattern of English. Triples were also used in \cite{das2011dynamic} to help extract events.

We can collect a set triples into a directed multigraph where the nodes correspond to the Head and Tail phrases of the triple and the edges correspond to the Relation phrase of the triple. Because multiple relations can exist between the same Head and Tail phrases, multiple edges can exist between a pair of nodes. In this paper, we will refer to such a graph constructed from semantic triples as an RDF (Resource Description Framework) graph due to the similarities in structure.

\subsection{Data Cleaning} \label{subsec: data clean}

Substantive efforts were put into cleaning the extracted triples. Some of the issues we encountered: 1) ambiguous pronoun references, 2) duplicate entities references such as \textit{United State} and \textit{America}. For the first problem, we used the co-reference tool NeuralCoref\footnote{https://github.com/huggingface/neuralcoref} provided in Spacy\footnote{https://github.com/explosion/spaCy}. For the second problem, we devised a multiple ways of computing similarity between phrases to detect semantic duplication. Table 1 shows the node phrase and edge phrase number, each phrase consist of multiple words. For example, \textit{thousands of people},\textit{Barack Obama's presidency}, To find duplicate phrases overall data, we need to compare each pair of phrases. This is a huge work. While such mature solutions (Spacy) can help us to deal with our data, how to efficiently process such big data is a challenge.


First, we determined duplication between phrases by counting the number of overlap words; we refer to this as coarse similarity. Let $P_1$ and $P_2$ be two phrases of interest. The coarse similarity distance between $P_1$ and $P_2$ is 
\[
\frac{|P_1 \cap P_2|}{\max(|P_1|, |P_2|)},
\]
where $|P_i|$ is the total number of words in the $i$th phrase and $|P_1 \cap P_2|$ is the number of words that occurred in both phrases. Thresholding the coarse similarity distance will help to associate phrases such as \textit{social media}, \textit{social medias posts}, and \textit{social medias platform} together. With the help of the first step, we cleaned majority duplicate phrases which have certain word in common, with less computation cost. The duplicate phrase defined by the coarse similarity distance is in Appendix \ref{A1:data detail}.

Next, we project the words in two different phrases into embedding space using Word2Vec\cite{mikolov2013distributed}; we refer to this as fine similarity. The fine similarity distance between $P_1$ and $P_2$ is 
\[
\frac{\sum\limits_{w_1\in P_1}^{}\sum\limits_{w_2 \in P_2}^{}\cos(w_1,w_2)}{|P_1|+|P_2|},
\]
where $\cos(w_1,w_2)$ is the cosine distance between the embedded vector representation of words in $P_1$ and $P_2$. Thresholding the fine similarity distance can help to connect related but seemingly dissimilar phrases such as \textit{emperor actor} and \textit{jussie smollett}. Reduction enable us to reduce the number of the triples from 25,734 to 5,045 in dataset JS and from 36,077 to 5,769 in dataset IP. The duplicate phrase defined by the fine similarity distance is in Appendix~\ref{A1:data detail}.



\section{Problem formulation}\label{ProDef}

Given a set of articles on a specific news event over time, $\mathcal{D} = \{D(1),\ldots,D(N)\}$, where $D(n)$ is the set of all the news articles collected on day $n$. Our goal is to develop a dynamic model of $D$ that includes 1) the semantic information contained in the articles and 2) the dynamic of news coverage (i.e., number of articles written on the topic per day). We transform the text data $D = \{D(1),\ldots,D(N)\}$ into a sequence of RDF graphs $\mathcal{G} = \{G(1),\ldots,G(N)\}$ consisting of semantic triples extracted from $D(n)$. Our goal is then to describe the dynamics of $\mathcal{G}$. 

Directly modeling the dynamics of a directed, a multigraph is mathematically difficult due to the combinatorial-sized state space. Instead, we can consider how substructures (i.e., triples) of $\mathcal{G}$ changes overtime. 

Let $\mathbf{G}_n^{n+m}$ denote the cumulative RDF graph from day $n$ to day $n+m$:
\[
\mathbf{G}_n^{n+m}  = G(n) \cup G({n+1}) \cup \ldots \cup G({n+m}).
\]
We can consider how a new graph (i.e. set triples) on day $n+1$, $G(n+1)$, may differ from the previous days graphs $\mathbf{G}_1^{n}$. In this paper, we consider three types of structural changes to $\mathbf{G}_1^{n}$:

\begin{description}
    \item[Append:] A new triple of $G({n+1})$ either appends to the head or tail of a triple in $\mathbf{G}_1^n$. New nodes and edges appear in $\mathbf{G}_1^{n+1}$ when append occurs. From a semantic perspective, occurrences of append events imply that new articles are presenting sentences/information that was unseen before. 
    
   
    \item[Extend:] A new triple of $G(n+1)$ connects two disconnected triples in $\mathbf{G}_1^n$. Graph $\mathbf{G}_1^{n+1}$ have additional edges and become more connected when extend occurs. From a semantic perspective, occurrences of extend events imply that connections are drawn between previous concepts. 
    

    
    \item[Mutate:] A new triple of $G({n+1})$ have the same head and tail as a triple in $\mathbf{G}_1^n$ but a different relationship. mutate events does not change the structure $\mathbf{G}_1^{n+1}$ if we discount multiple edges. From a semantic perspective, occurrences of mutate events imply that alternative interpretation/wordings are presented. 
    


\end{description}
\begin{figure}
  \centering
  \includegraphics[scale=0.28]{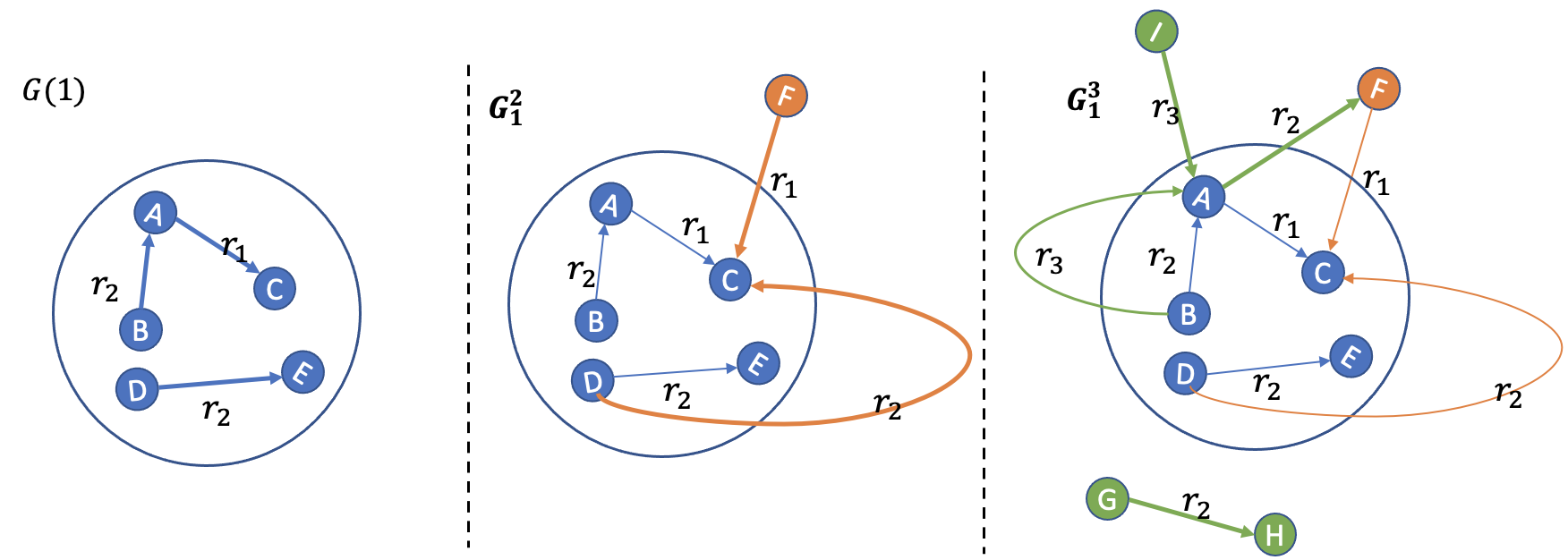}
  \caption{Example RDF Graph Sequence (Red = New Triples on Day 2, Green = New Triples on Day 3)}
\label{dynamicdata}
\end{figure}

Figure~\ref{dynamicdata} shows an example illustrating how the cumulative RDF graphs grows from $G(1)$ to $\mathbf{G}_1^3$. The initial set of triples in $G(1)$ are $A \rightarrow r_1 \rightarrow C$, $B \rightarrow r_2 \rightarrow A$, and $D \rightarrow r_2 \rightarrow E$. We will track what happens to these initial triples over time.

\begin{itemize}
\item On day 2, the new triple $F \rightarrow r_1 \rightarrow C$ appends to the initial triple $A \rightarrow r_1 \rightarrow C$ and the new triple $D \rightarrow r_2 \rightarrow C$ is an extension of $A \rightarrow r_1 \rightarrow C$. On day 3, the new triple $I \rightarrow r_3 \rightarrow A$ appends to $A \rightarrow r_1 \rightarrow C$. Two new triples, $A \rightarrow r_2 \rightarrow F$, $B \rightarrow r_3 \rightarrow A$, are extensions of $A \rightarrow r_1 \rightarrow C$ 

\item On day 2, the initial triple $B \rightarrow r_2 \rightarrow A$ is unchanged. On day 3, the new triple $B \rightarrow r_3 \rightarrow A$ is a mutation of $B \rightarrow r_2 \rightarrow A$, $I \rightarrow r_3 \rightarrow A$ appends to $B \rightarrow r_2 \rightarrow A$ and $A \rightarrow r_2 \rightarrow F$ are extensions of $B \rightarrow r_2 \rightarrow A$.

\item On day 2, the new triple $D \rightarrow r_2 \rightarrow C$ is an extension of the initial triple $D \rightarrow r_2 \rightarrow E$. On day 3, $D \rightarrow r_2 \rightarrow E$ is unchanged.
\end{itemize}

By keeping track of the append, extend, mutate events, we can characterize how the cumulative RDF graph changes over time. Let $\mathcal{I}$ denote the initial set of triples. Given $\mathcal{I}$, each initial triple induces a multivariate count time series over time:
\begin{equation}\label{eq:timseries}
\mathbf{y}^{(i)}(n) = \begin{bmatrix}
y^{(i)}_{\text{append}}(n)  \\
y^{(i)}_{\text{extend}}(n)\\
y^{(i)}_{\text{mutate}}(n)
\end{bmatrix}, i \in \mathcal{I}, n =1, \ldots, N, 
\end{equation}
where $y^{(i)}_{\text{append}}(n)$ is the total number of new triples on day $n$ that are appended to triple $i$; $y^{(i)}_{\text{extend}}(n)$ is the total number of new triples on day $n$ that are extension of $i$, and $y^{(i)}_{\text{mutate}}(n)$ is the total number of triples on day $n$ that are mutations of $i$. 

The problem of characterizing the evolution of $D(n)$ by way of accounting for the structural changes of the cumulative RDF graph $\mathbf{G}_{1}^N$ can be done by fitting time series models to $\mathbf{y}^{(1)}(n)$, $\mathbf{y}^{(2)}(n),\ldots,\mathbf{y}^{(|\mathcal{I}|)}(n)$. 

\section{Discrete-Time Hawkes Process}\label{sec:DHP}

The classic Hawkes process is a continuous-time point model. This means that event cannot occur simultaneously; this is a reasonable assumption for a continuous-time system. However, our data was collected per day. Therefore, it is very likely that multiple events occurred on the same day. As a result, the process is no longer a point process but a counting process. In this paper, we used the discrete-time variation of the Hawkes process, introduced in \cite{hawkes2021discrete} to model multivariate count time series. 

Consider a discrete-time Hawkes process, $\mathbf{y}(n)=[y_1(n),\ldots, y_M(n)]^T$, where $y_m(n)$ represents the number of occurrences of the $m$th type of event in the $n$th time interval. Let $\mathbf{H}_1^{n-1} = \{\mathbf{y}(1), \ldots, \mathbf{y}(n-1)\}$ denote the history of the process up to time interval $n-1$. The discrete-time Hawkes process is characterized by the conditional intensity function
\begin{equation}\label{eq:rate}
\lambda(n) = E[\mathbf{y}(n)|\mathbf{H}_1^{n-1}] = \mathbf{\mu}+\sum_{t<n}\mathbf{A} \mathbf{y}(t)\phi(n-t),
\end{equation}
where $\mu = [\mu_{1},\ldots, \mu_{M}]^T$ is the baseline vector. And $A$ is the $M \times M$ infectivity matrix where $A_{ij}$ shows how the $i$th event type is influenced by the $j$th event type. The function $\phi(t)$ is the delay function. A popular delay function is the exponential function: 
\begin{equation}\label{eq:exp}
    \phi(t) := \beta e^{-\beta t}, t\ge 0.
\end{equation}\label{equ:phi}
Each time-series $y_m(n)$ is a Poisson process with rate $\lambda_m(n)$.
\begin{equation}
    P(y_m(n);\lambda_m(n)) = \frac{(\lambda_m(n))^{y_m(n)}}{y_m(n)!}e^{-\lambda_m(n)}, \quad 1 \le m \le M.
\label{equ:poisson}
\end{equation}
Even though $P(y_m(n);\lambda_m(n))$ is not explicitly a conditional distribution where $y_m(n)$ depends on the past values $y_m(n-1), y_m(n-2), \ldots$, dependence on past values is captured implicitly through $\lambda_m(n)$.

\subsection{Parameter Estimation}

The parameters of the discrete-time, multivariate Hawkes process are $\Theta = \{\mu, A\}$. The parameter $\beta$ of the delay function is a hyperparameter. We will estimate $\Theta$ using maximum likelihood:
\[
\widehat{\Theta} = \arg \max_{\Theta} \mathcal{L}(\Theta),
\]
where the log-likelihood function is
\begin{align*}
     \mathcal{L}(\Theta) & = \log\left(\prod_{m = 1}^{M}\prod_{n = 1}^{N} P(y_m(n);\lambda_m(n))\right)\\
     & = \sum_{m=1}^M\sum_{n = 1}^{N} y_m(n)\log(\lambda_m(n))-\lambda_m(n)-log(y_m(n)!).
\end{align*}

\section{Experiments}\label{sec:exp}

Given an initial set of triples, $\mathcal{I}$, we constructed sets of induced multivariate count time series, $\mathbf{y}^{(i)}(n), i \in \mathcal{I}$, from our collected text data as described in Section~\ref{ProDef}. In this section, we will show that $\mathbf{y}^{(i)}(n)$ reflects both the dynamics of how many articles were published each day as well as the semantic content of those articles. By modeling $\mathbf{y}^{(i)}(n)$, we can extract information from the system that we can not do by only looking at article count dynamics nor by only considering the text in the articles.




\subsection{Major vs. Entertainment News Outlet Dynamics}

From the set of collected new articles $\mathcal{D}$, we extracted $\mathcal{D}_{\text{Major}}$ , which contains articles from major news outlets\footnote{\url{https://blog.feedspot.com/usa_news_websites/}} such as Politico, BBC, New York Times, NPR, Reuters, Mercury News, etc., and $\mathcal{D}_{\text{Ent}}$, which contains articles from entertainment outlets \footnote{\url{https://blog.feedspot.com/celebrity_gossip_blogs/};\url{https://aelieve.com/rankings/websites/category/news-media/top-celebrity-gossip-websites/};\url{https://blog.feedspot.com/pop_culture_blogs/}} such as TMZ, People, E!Online, GlobalNews, The Sun, Pop Sugar, etc (see Appendix~\ref{A3: news outlets} for examples).

\begin{figure*}[ht]
     \centering
     \hfill
     \begin{subfigure}{0.45\textwidth}
         \centering
         \includegraphics[width=1.0\textwidth]{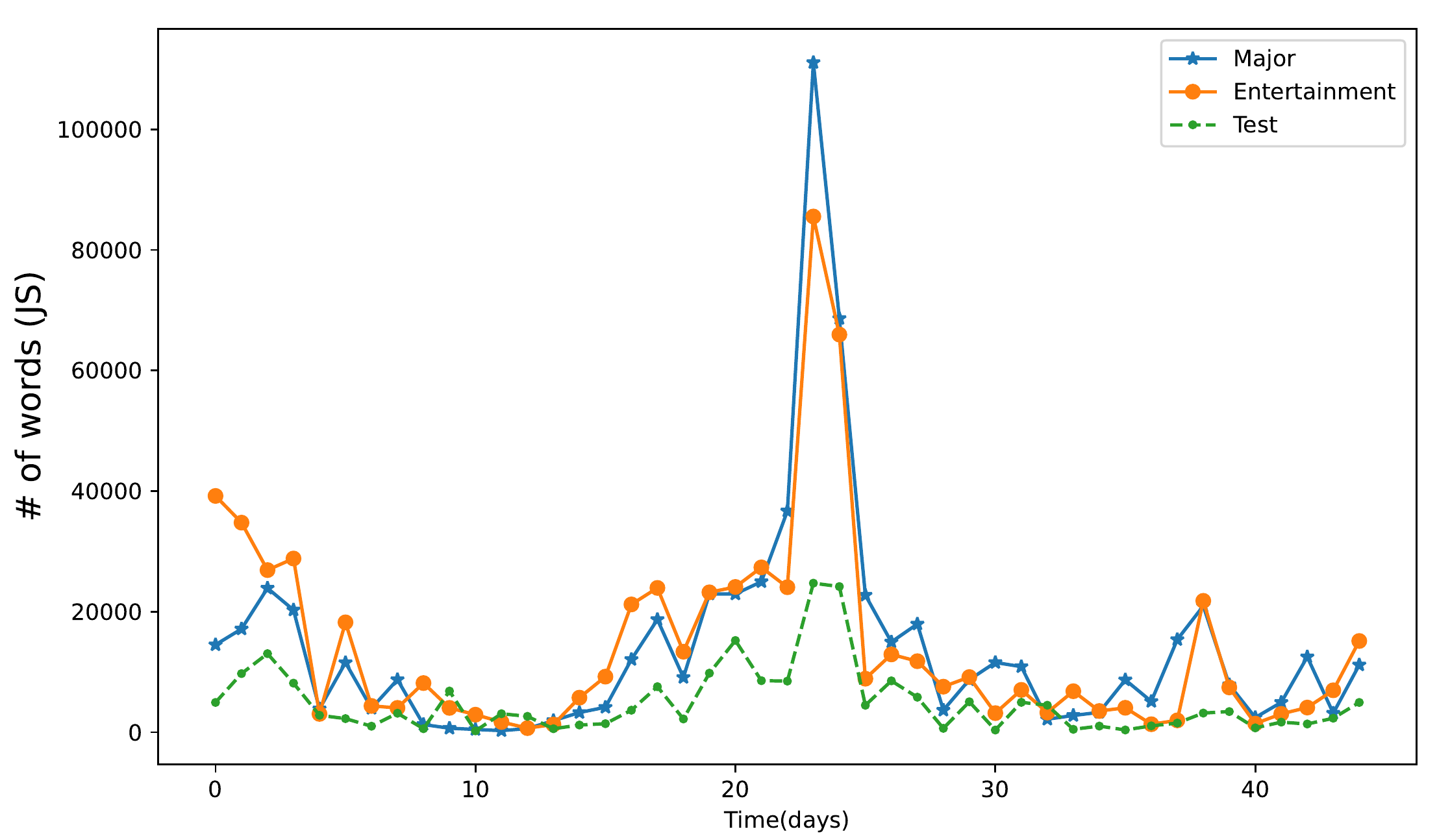}
         \caption{Word Count of Dataset JS }
         \label{JS Bignew3s}
     \end{subfigure}
     \hfill
     \begin{subfigure}{0.45\textwidth}
         \centering
         \includegraphics[width=1.0\textwidth]{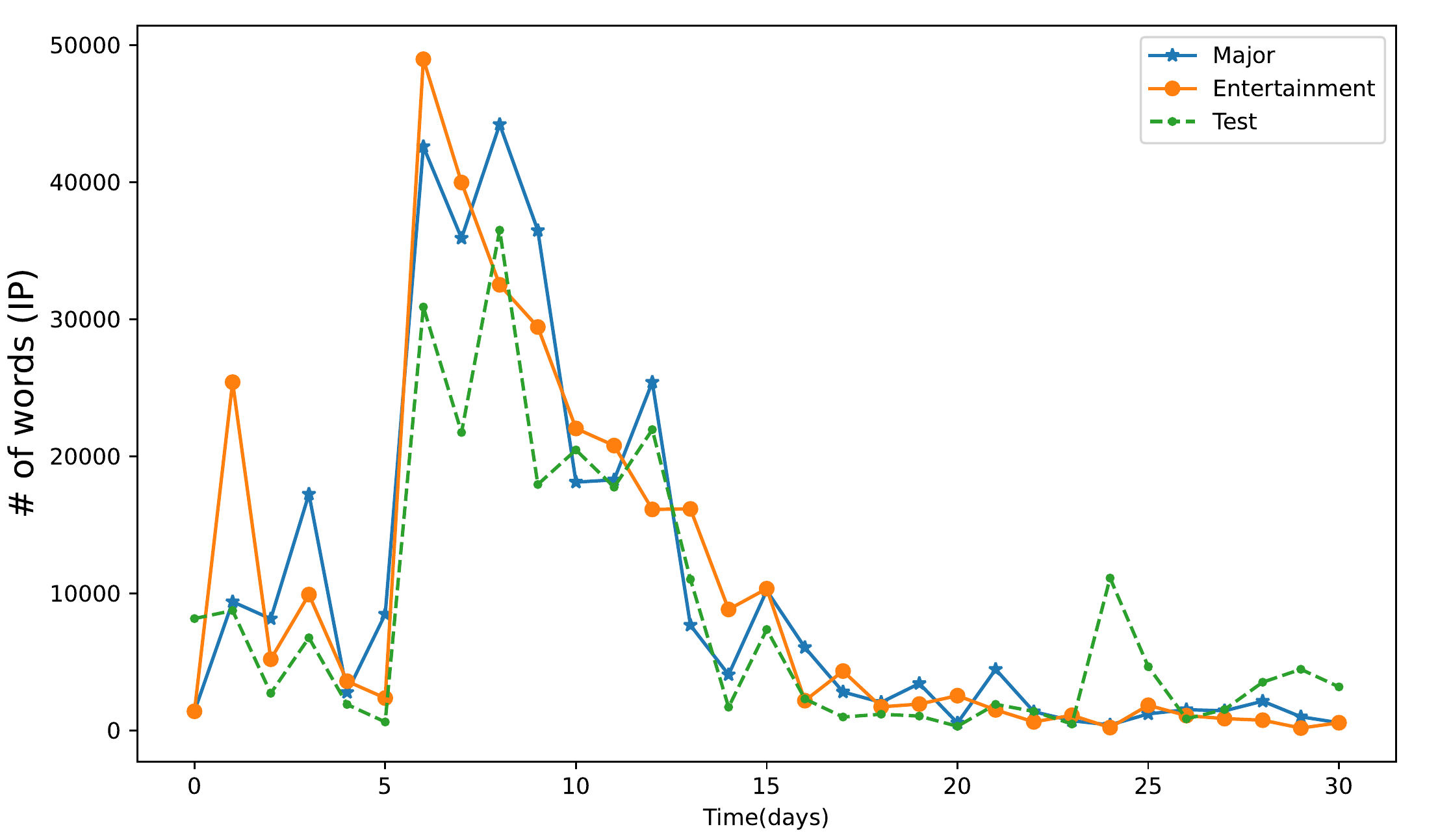}
         \caption{The words count of dataset IP }
         \label{IP Gossip3}
     \end{subfigure}
     \hfill
        \caption{Word Count of Major and Entertainment News Outlets}.
        \label{count of BG1}
\end{figure*}

We specified a common set of triples $\mathcal{I}$ (e.g.jussie smollett$\rightarrow$play$\rightarrow$ jamal lyon, ). Because the number of news outlets and articles is so large, we considered a small number initial triples collected broadly from news articles collected from the first few days of coverage. For dataset JS, $|\mathcal{I}| = 16$. For dataset IP, $|\mathcal{I}| = 42$ (see Appendix~\ref{A2: common triple} for examples). The initial triples induce two sets of multivariate count time series, $\mathbf{y}^{(i)}_{\text{Major}}(n), i \in \mathcal{I}$ from $\mathcal{D}_{\text{Major}}$ and $\mathbf{y}^{(i)}_{\text{Ent}}(n), i \in \mathcal{I}$ from $\mathcal{D}_{\text{Ent}}$. We normalized counts of append, extend, and mutate events by their respective averages so they can be considered at the same scale. 

\subsection{Model Fitting}

We fit $\mathbf{y}^{(i)}_{\text{Major}}(n), i \in \mathcal{I}$ and $\mathbf{y}^{(i)}_{\text{Ent}}(n), i \in \mathcal{I}$ to the discrete-time Hawkes process introduced in Section~\ref{sec:DHP}. This will inform us how the initial triples, $||$ and $||$, change with append, extend, and mutate. We treated the time series induced by the $i$th triple as independent observations and solved for the maximum likelihood estimator, $\widehat{\Theta}$, which included the $3\times 1$ baseline rate vector, $\widehat{\mu} = [\widehat{\mu}_{\text{append}}, \widehat{\mu}_{\text{extend}}, \widehat{\mu}_{\text{mutate}}]^T$ and the $3 \times 3$ infectivity matrix, $\widehat{\mathbf{A}}$. We set the hyperparameter $\beta$ in~\eqref{eq:exp} to 0.5.

\begin{figure}[ht]
     \centering
     \hfill
     \begin{subfigure}{0.23\textwidth}
         \centering
         \includegraphics[width=1.0\textwidth]{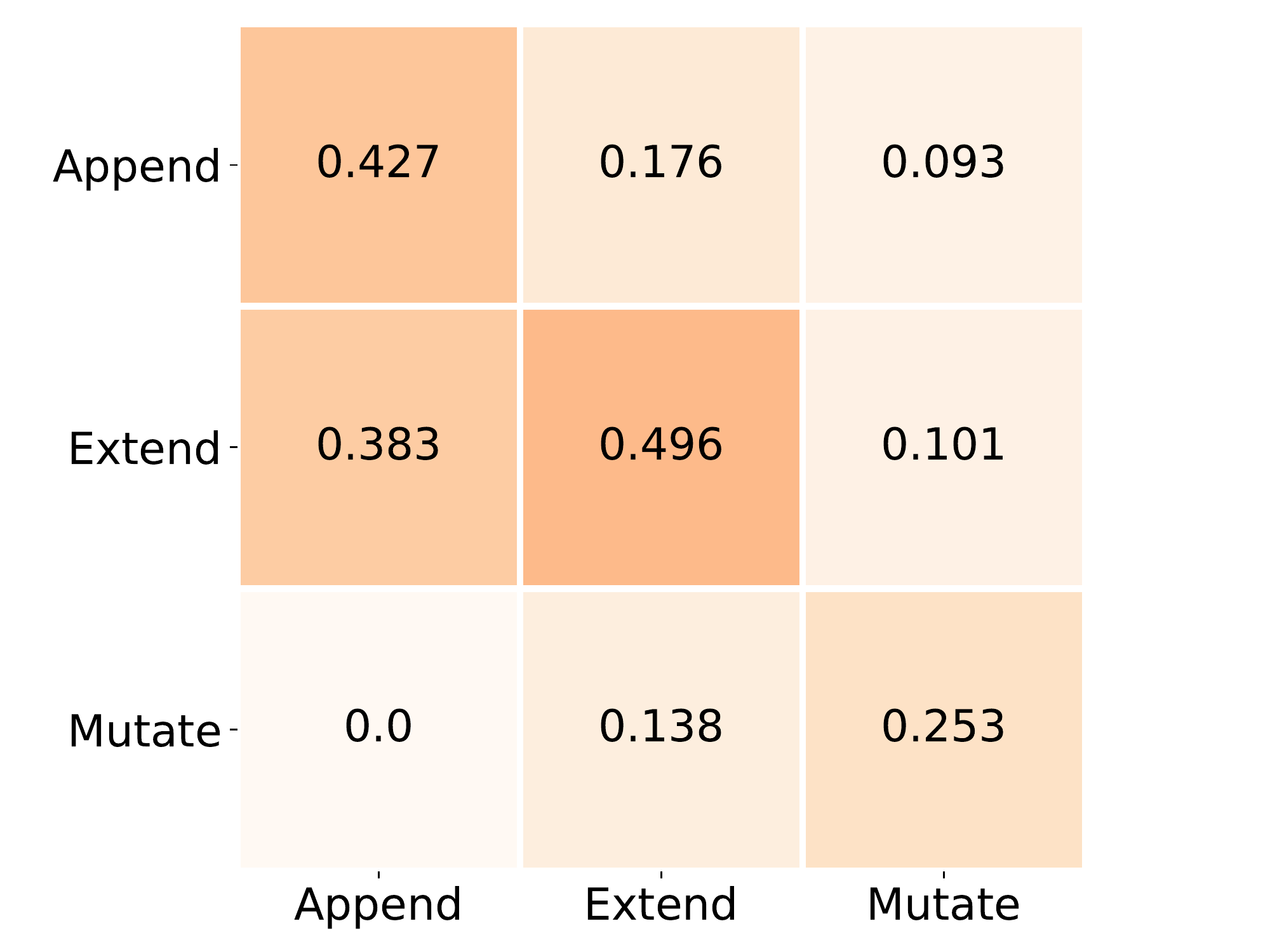}
         \caption{Major News Outlets (JS) }
         \label{JS Bignews2}
     \end{subfigure}
     \hfill
     \begin{subfigure}{0.23\textwidth}
         \centering
         \includegraphics[width=1.0\textwidth]{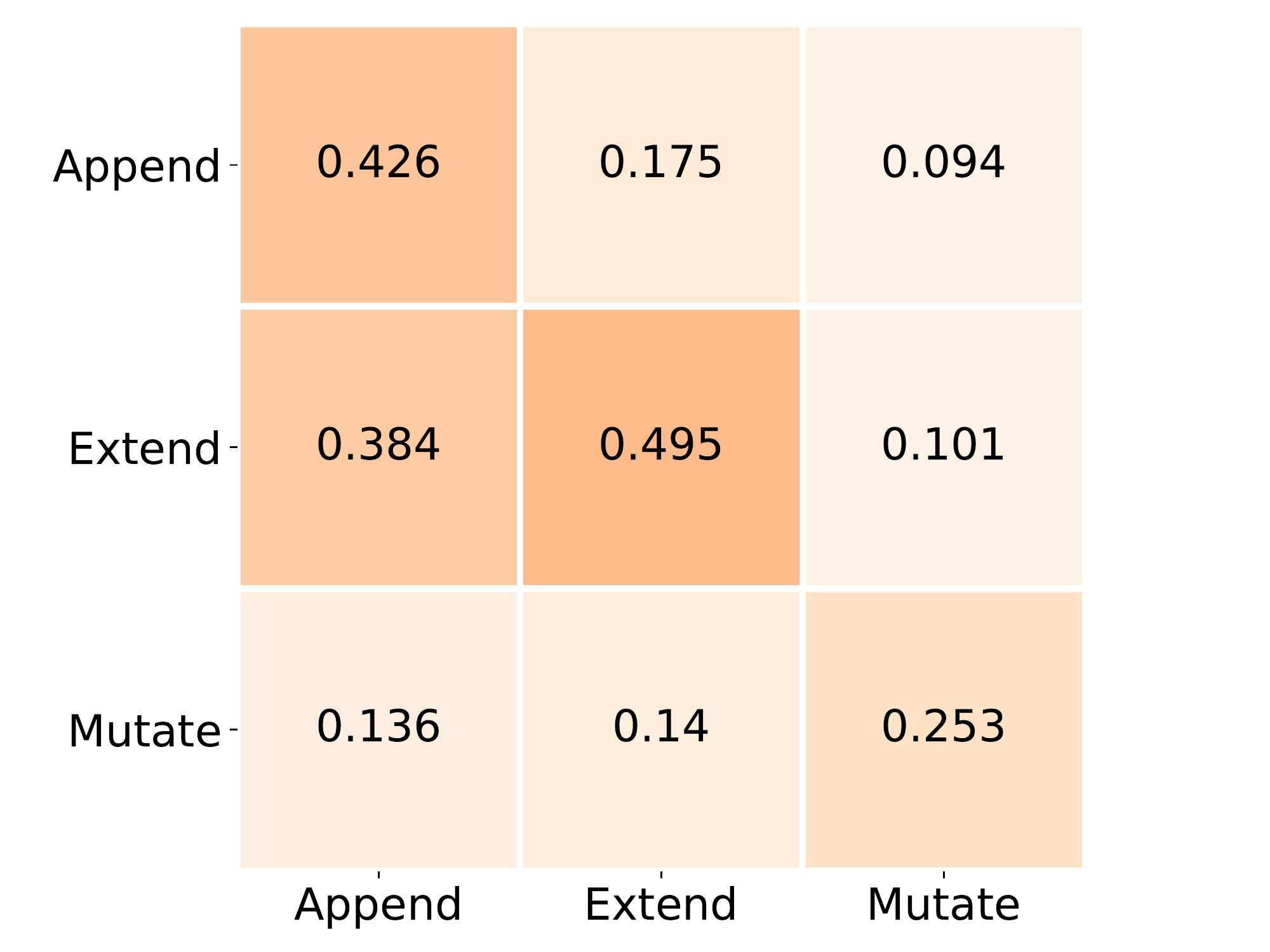}
         \caption{Entertainment News Outlets (JS)}
         \label{JS Gossip}
     \end{subfigure}
     \begin{subfigure}{0.23\textwidth}
         \centering
         \includegraphics[width=1.0\textwidth]{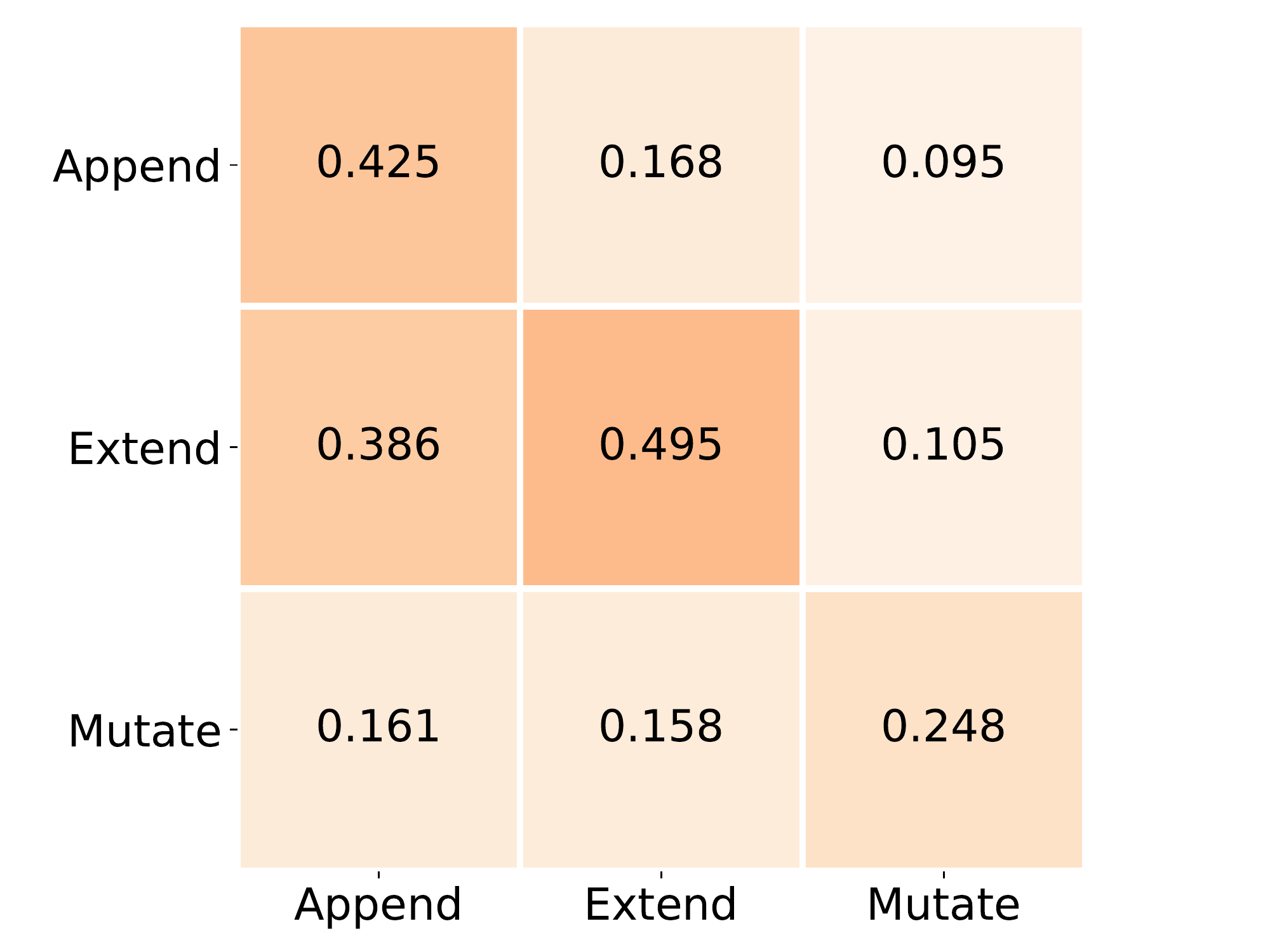}
         \caption{Major News Outlets (IP) }
         \label{IP Bignews}
     \end{subfigure}
     \hfill
     \begin{subfigure}{0.23\textwidth}
         \centering
         \includegraphics[width=1.0\textwidth]{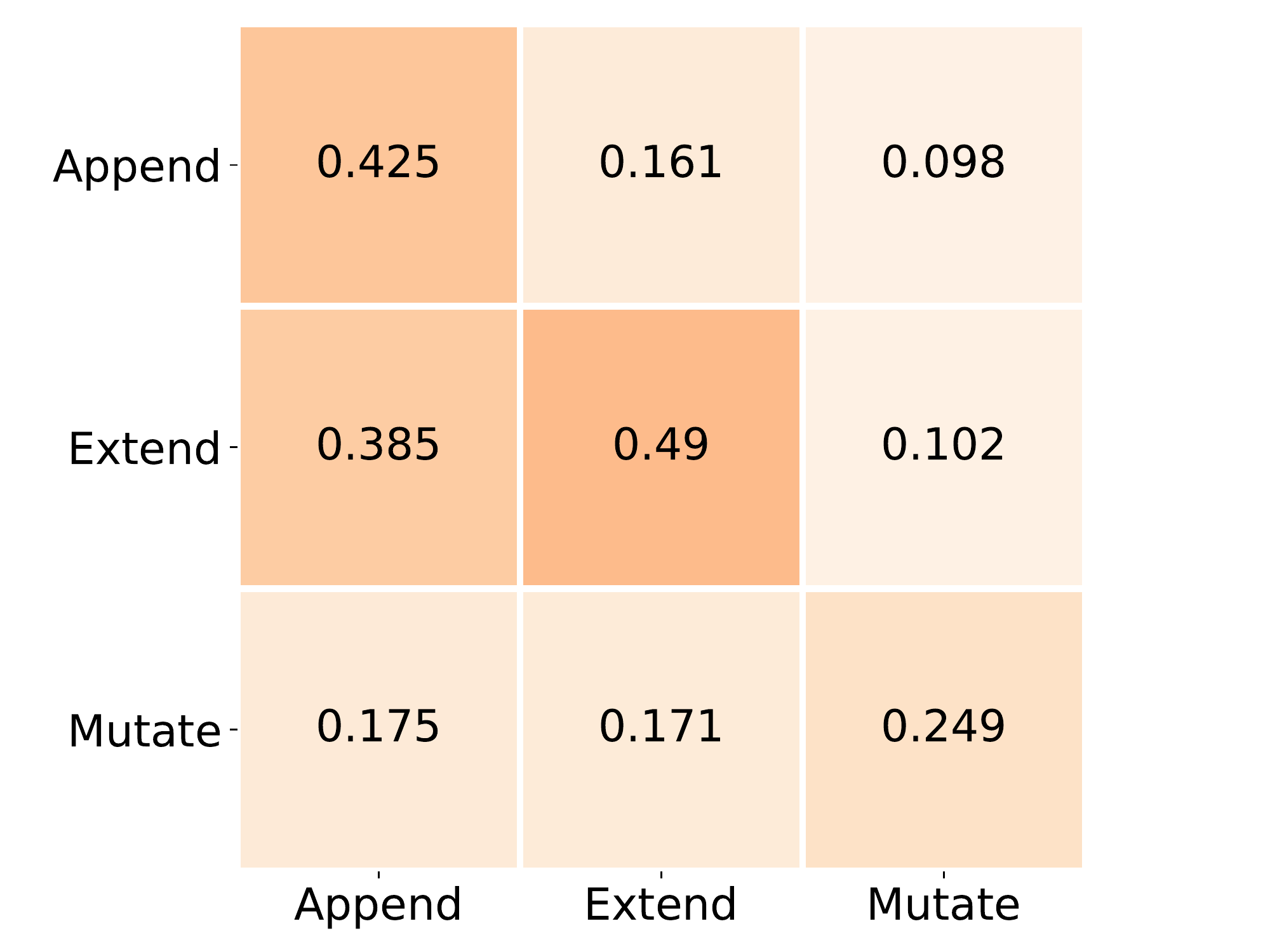}
         \caption{Entertainment News Outlets (IP) }
         \label{IP Gossip2}
     \end{subfigure}
     \hfill
        \caption{Infectivity Matrix}
        \label{mutate infection}
\end{figure}

Let $\widehat{\mu}_{\text{Major}}$ and $\widehat{\mu}_{\text{Ent}}$ denote the baseline rate vector and $\widehat{\mathbf{A}}_{\text{Major}}$ and $\widehat{\mathbf{A}}_{\text{Ent}}$ denote the infectivity matrix estimated from $\mathbf{y}^{(i)}_{\text{Major}}(n), i \in \mathcal{I}$ and $\mathbf{y}^{(i)}_{\text{Ent}}(n), i \in \mathcal{I}$, respectively. Table~\ref{BigGossip mu} shows the baseline rate vector estimates and Figure~\ref{mutate infection} shows the infectivity matrix estimates.

\begin{table}[ht]
\centering
  \caption{Baseline Rate Vector Estimates}
  \label{BigGossip mu}
\begin{tabular}{c|c|c|c}
 \hline
Dataset  & & $\widehat{\mu}_{\text{Major}}$ & $\widehat{\mu}_{\text{Ent}}$\\
 \hline
JS   & Append  & 0.3904 &0.5172\\
    & Extend  &0.3011 &0.3023 \\
  &Mutate   &0.5701 &0.7924 \\

 \hline
IP   & Append  & 0.5867 &0.5905\\
    & Extend  &0.2602 &0.2646 \\
  &Mutate   &0.4591 &0.6139 \\
 \hline
\end{tabular}
\end{table}

We can see that news articles published by entertainment outlets have a higher baseline rate for append and mutate than news articles published by major news outlets. Since the count of append reflects how completely new information is added to an initial triple while mutate reflects re-wording/re-interpretation of an initial triple. This reflects potentially that entertainment news articles have more diverse coverage pertaining to the subject of Jussie Smollet and Flight 752 than major news outlets. 

Similarly from the infectivity matrices, we see that mutate events from articles published by entertainment outlets induce more append and extend events than major news outlet articles.

\begin{figure*}[htb]
     \centering
     \begin{subfigure}{0.45\textwidth}
         \centering
         \includegraphics[width=1.0\textwidth]{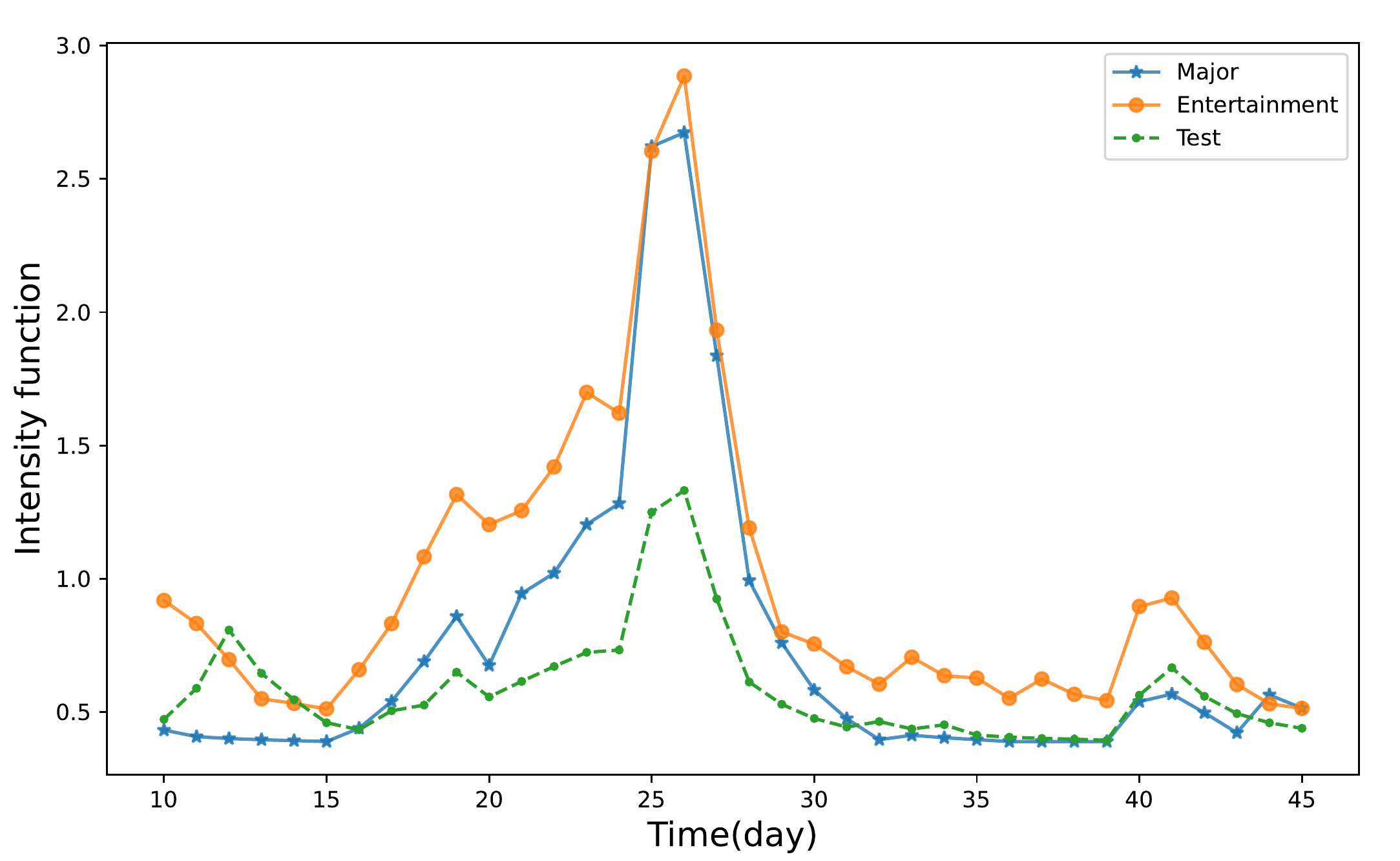}
         \caption{ $\overline{\lambda_{\text{append}}}(n)$ of dataset JS}
         \label{JS append lambda classify}
     \end{subfigure}
     \begin{subfigure}{0.45\textwidth}
         \centering
         \includegraphics[width=1.0\textwidth]{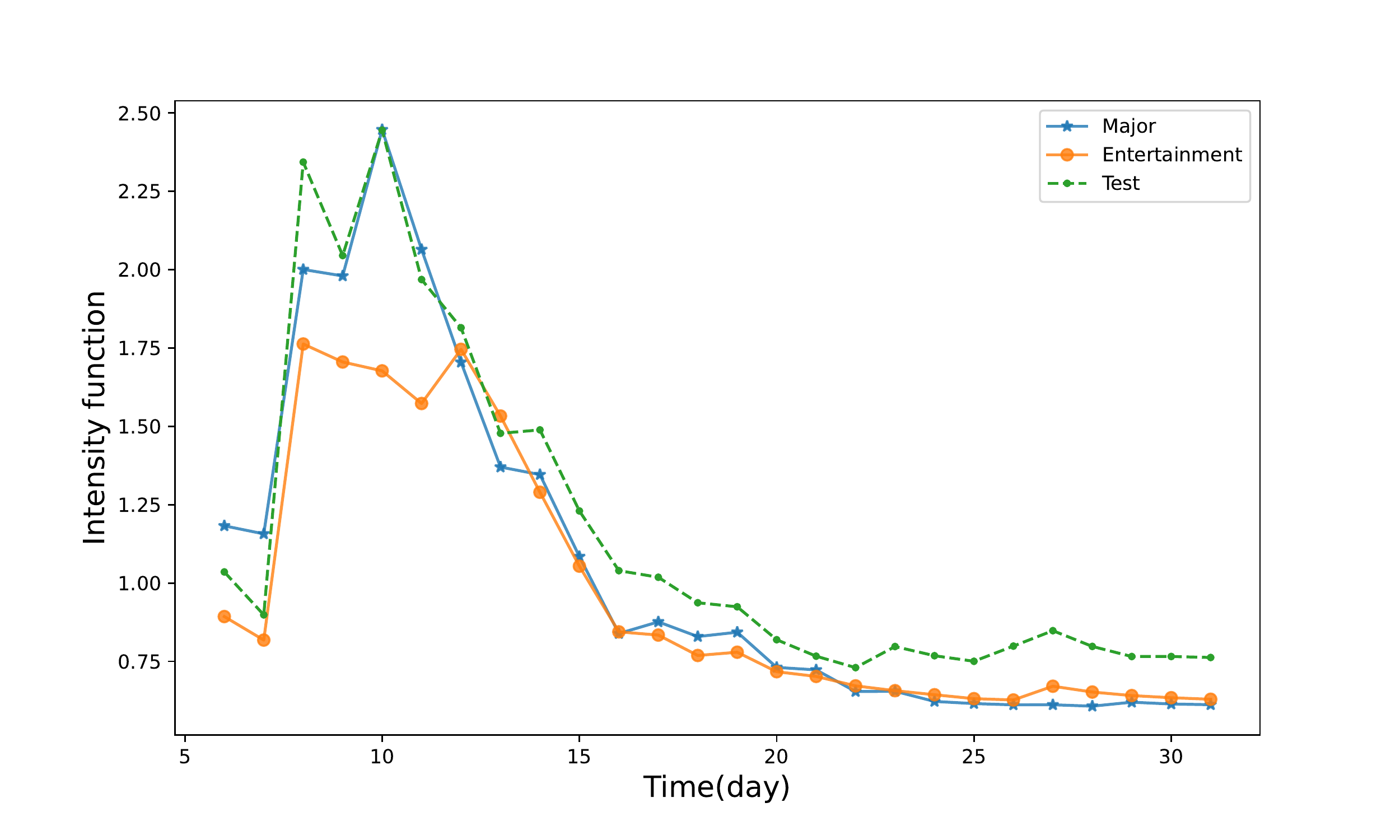}
         \caption{ $\overline{\lambda_{\text{append}}}(n)$ of dataset IP}
         \label{IP append lambda classify}
     \end{subfigure}
     
     \begin{subfigure}{0.45\textwidth}
         \centering
         \includegraphics[width=1.0\textwidth]{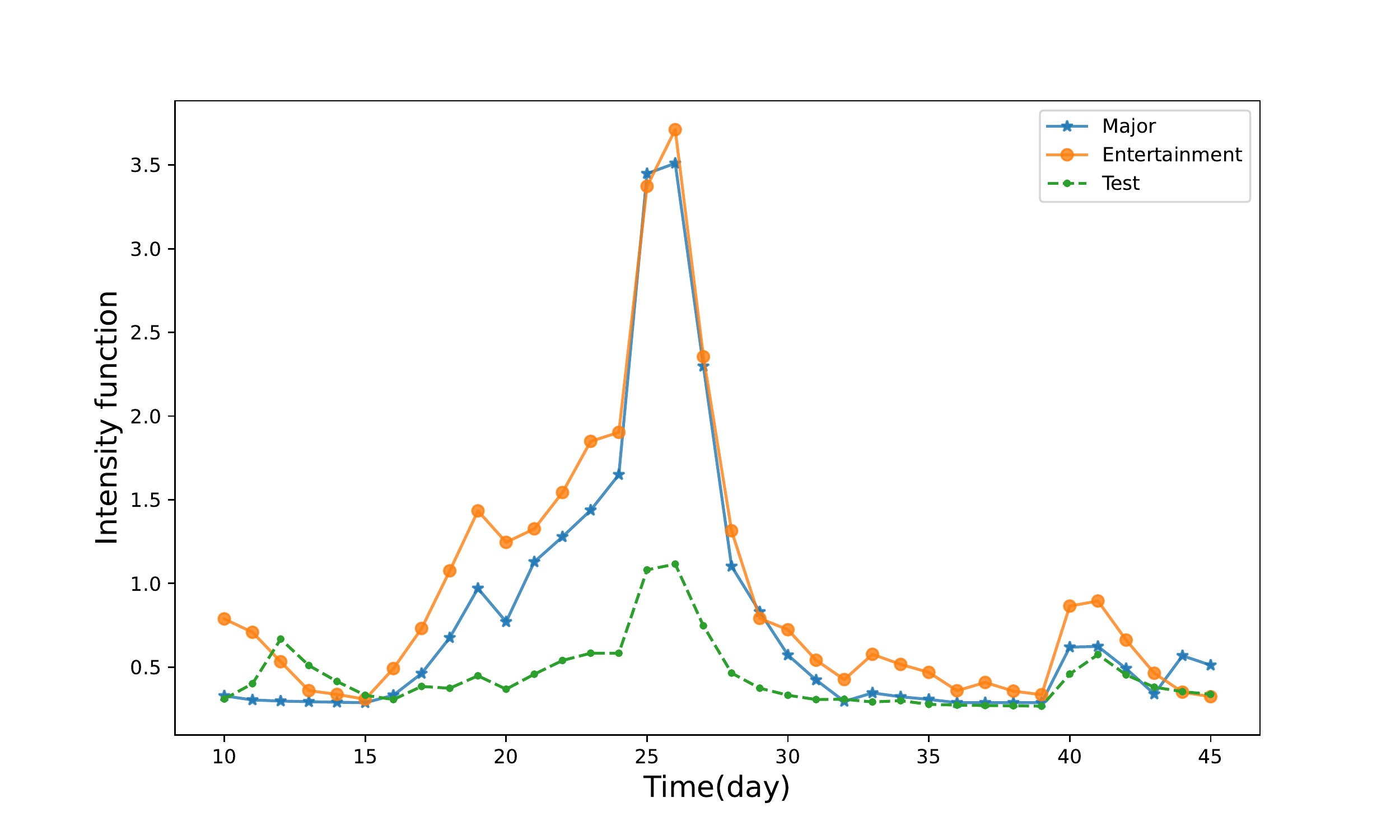}
         \caption{$\overline{\lambda_{\text{Extend}}}(n)$  of dataset JS}
         \label{JS extend lambda classify}
     \end{subfigure}
     \begin{subfigure}{0.45\textwidth}
         \centering
         \includegraphics[width=1.0\textwidth]{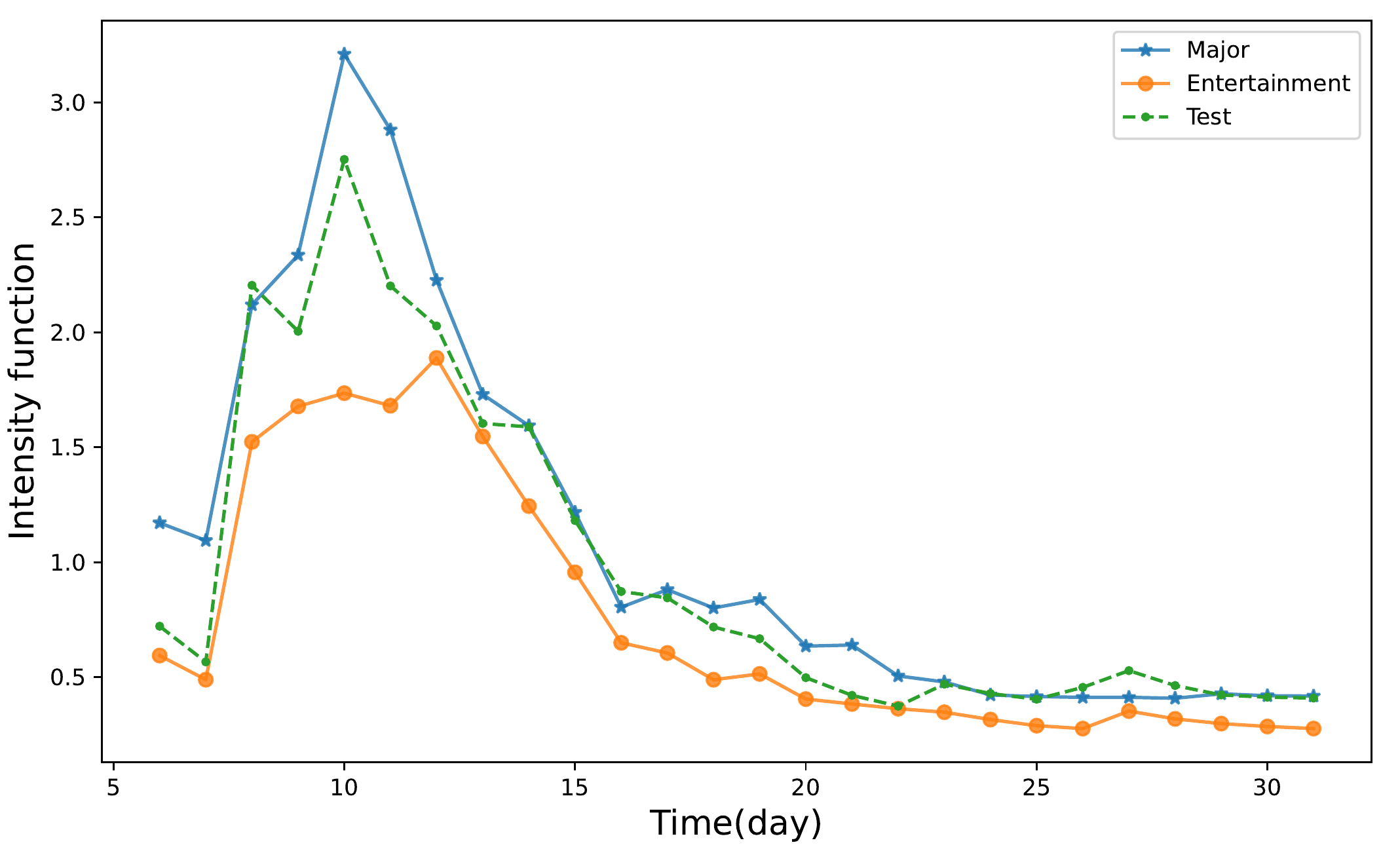}
         \caption{$\overline{\lambda_{\text{extend}}}(n)$  of dataset IP}
         \label{IP extend lambda classify}
     \end{subfigure}
     \begin{subfigure}{0.45\textwidth}
         \centering
         \includegraphics[width=1.0\textwidth]{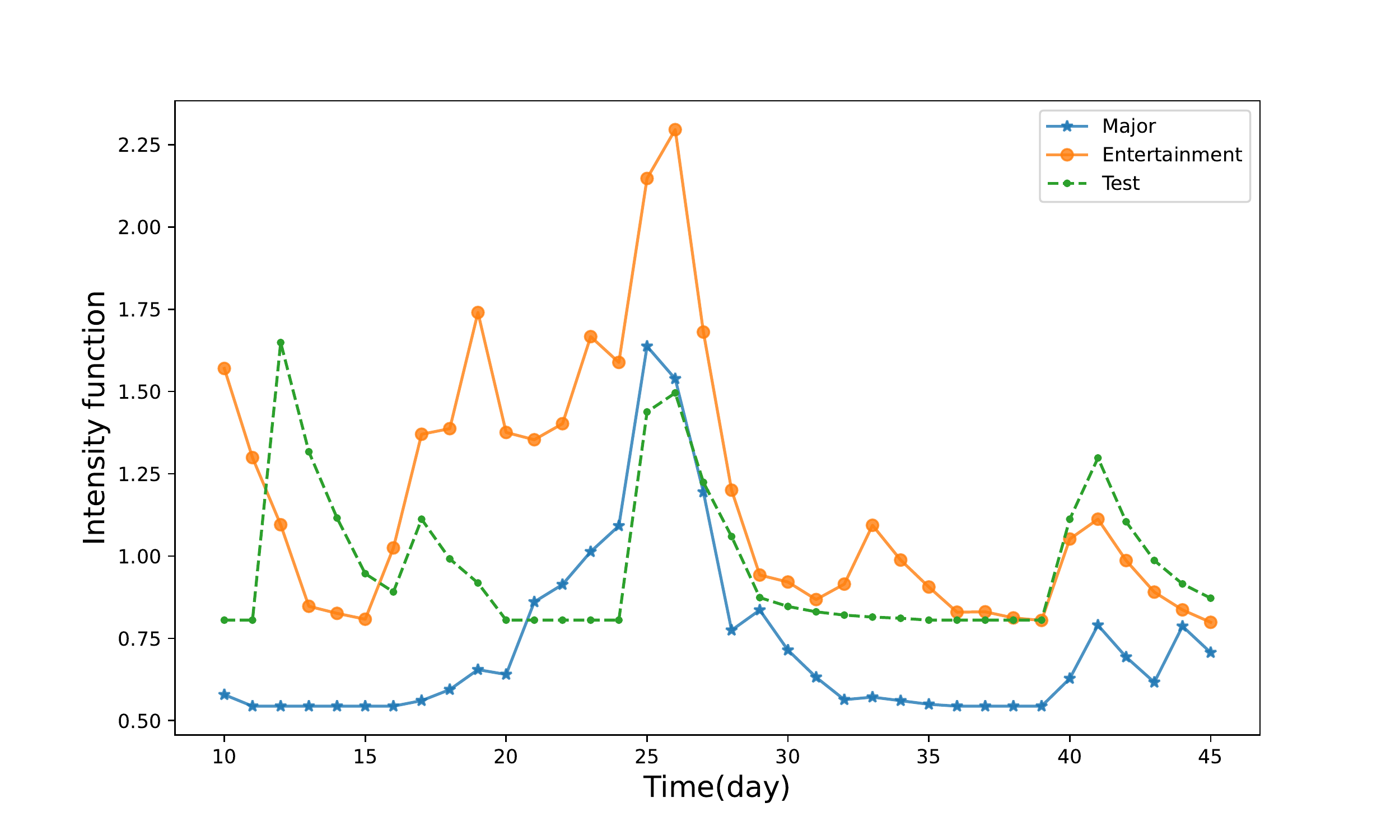}
         \caption{$\overline{\lambda_{\text{Mutate}}}(n)$  of dataset JS}
         \label{JS mutate lambda classify}
     \end{subfigure}
     \begin{subfigure}{0.45\textwidth}
         \centering
         \includegraphics[width=1.0\textwidth]{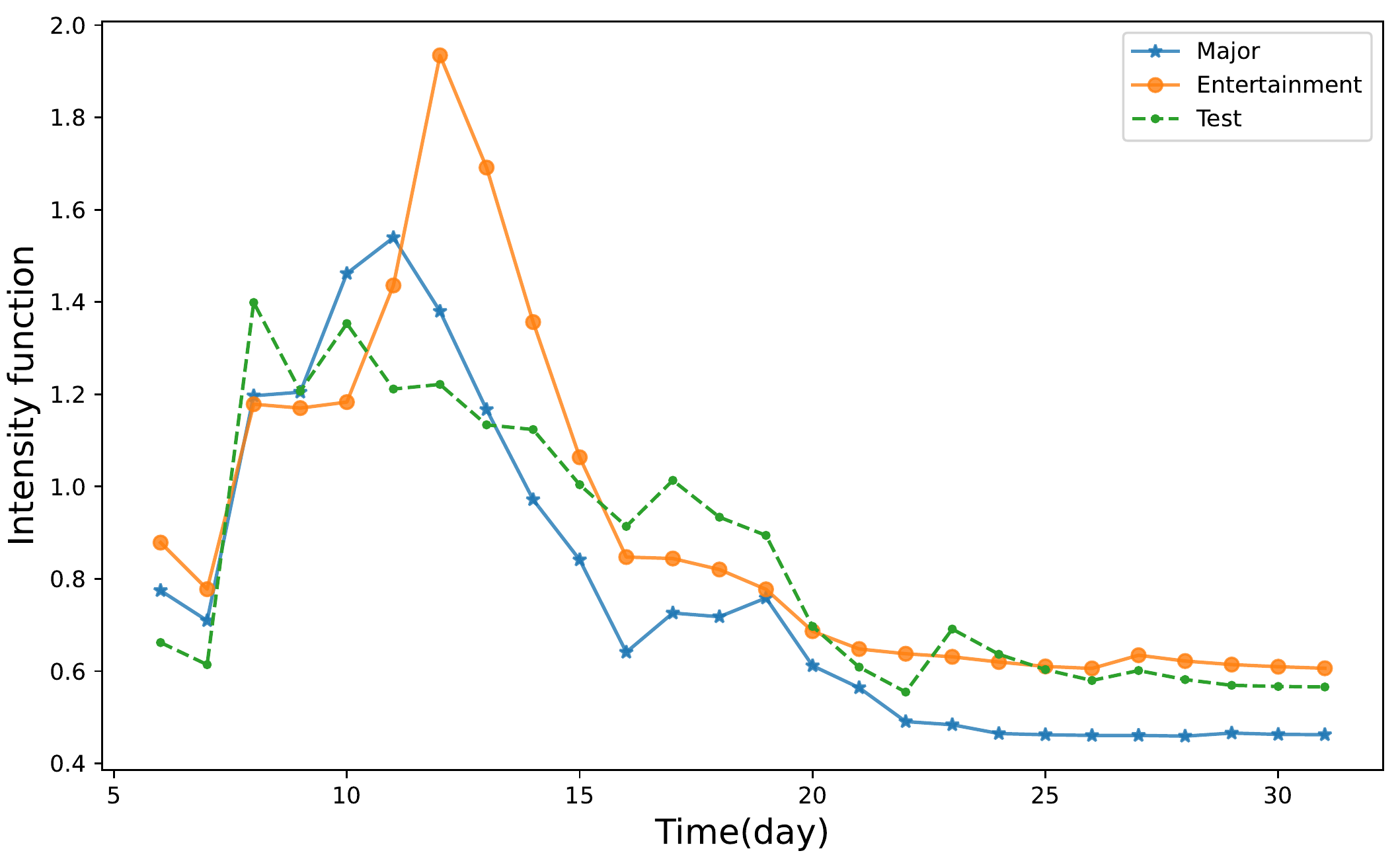}
         \caption{$\overline{\lambda_{\text{mutate}}}(n)$  of dataset IP}
         \label{IP mutate lambda classify}
     \end{subfigure}
        \caption{Average Conditional Intensity Function $\overline{\lambda}(n)$}
        \label{lambda classify}
\end{figure*}

We used the estimated parameters to find the conditional intensity function \eqref{eq:rate}, $\widehat{\lambda}_{\text{append}}^{(i)}(n)$, $\widehat{\lambda}_{\text{extend}}^{(i)}(n), \widehat{\lambda}_{\text{mutate}}^{(i)}(n), i \in \mathcal{I}$, which is the expected number of occurrences of append, extend, and mutate given past observations. Figure~\ref{lambda classify} plots the average conditional intensity
\[
\overline{\lambda_m(n)} = \frac{1}{|\mathcal{I}|}\sum_{i \in \mathcal{I}} \widehat{\lambda}_m^{(i)}(n), m \in \{\text{append, extend, mutate}\}
\]
for $\mathbf{y}^{(i)}_{\text{Major}}(n)$ and $\mathbf{y}^{(i)}_{\text{Ent}}(n)$ in dataset JS and IP. 

The average conditional intensity function captures the dynamics of article posts as the peak of append, extend, and mutate events coincide to days when the number of articles also peaked (see Figure~\ref{article of total}). However, since the count of append, extend, and mutate are induced by semantic information within the article, they capture the difference in information content/writing style in articles from major and entertainment news articles.

\subsection{Classifying Unknown News Sources}

Consider a set of news articles covering the same events but from an unknown news source. We will call this the test set, $\mathcal{D}_{\text{Test}}$. Since we have obtained models for major and entertainment news outlets, we will see if we can determine if the outlets that generated $\mathcal{D}_{\text{Test}}$ can be considered to be major news outlets or entertainment news outlets.

For our experiment, the test sets for the dataset JS and dataset IP are comprised of 10 known news sources such as CNN, USA Today, Breitbart, Newsweek, etc. The list of news outlets can be found in Appendix~\ref{A3: news outlets}. Due to our choice of news outlet for the test set, we expect $\mathcal{D}_{\text{Test}}$ have similar coverage dynamics as $\mathcal{D}_{\text{Major}}$. Note that $\mathcal{D}_{\text{Test}}$ comprised of fewer articles than $\mathcal{D}_{\text{Major}}$ and $\mathcal{D}_{\text{Ent}}$. Figure~\ref{count of BG1} shows the total number of words each day in the different datasets.

A set of multivariate count time series, $\mathbf{y}^{(i)}_{\text{Test}}(n)$, is generated using the same initial triples $\mathcal{I}$ that generated $\mathbf{y}^{(i)}_{\text{Major}}(n)$ and $\mathbf{y}^{(i)}_{\text{Ent}}(n)$. The discrete-time Hawkes process is fitted to the data and the average condition intensity function for $\mathbf{y}^{(i)}_{\text{Test}}(n)$ is plotted in Figure \ref{lambda classify}. 

Table~\ref{predition} shows the L1 and L2 distance between the average conditional intensity functions of the test set, $\overline{\lambda(n)}_{\text{Test}}$, with $\overline{\lambda(n)}_{\text{Major}}$ and $\overline{\lambda(n)}_{\text{Ent}}$. The distance between the test outlets and the major news outlets using the L2 norm is always the smallest for append, extend, and mutate events. This means that $\mathcal{D}_{\text{Test}}$ is the most similar to $\mathcal{D}_{\text{Major}}$ as we expected. 

\begin{table}
\centering
  \caption{Average Conditional Intensity Function Distance for ($\mathcal{D}_{\text{Test}}$, $\mathcal{D}_{\text{Major}}$) and ($\mathcal{D}_{\text{Test}}$, $\mathcal{D}_{\text{Ent}}$)}
  \label{predition}
\begin{tabular}{p{1cm}|p{1cm}p{1cm}||cc}
 \hline
\multirow{2}{*}{Dataset} & \multirow{2}{*}{}& {News } &L1 &L2  \\
{} & &outlets& norm&norm\\
 \hline

\multirow{8}{*}{JS}   & \multirow{2}{*}{Append}& Major &\textbf{0.2320} & \textbf{0.1641} \\
& &Ent &0.4012 & 0.2959 \\
\cline{2-5}
& \multirow{2}{*}{Extend}& Major &\textbf{0.3933} & \textbf{0.5042}   \\
& &Ent &0.5371 & 0.6187   \\
\cline{2-5}
& \multirow{2}{*}{Mutate}& Major & \textbf{0.2990} &\textbf{0.1332}   \\
& &Ent &0.3134 & 0.1748 \\

\hline
\hline
\multirow{8}{*}{IP}   & \multirow{2}{*}{Append}& Major &\textbf{0.1401} & \textbf{0.0243}  \\
& &Ent &0.1920& 0.0625 \\
\cline{2-5}
& \multirow{2}{*}{Extend}& Major &\textbf{0.1540} & \textbf{0.0565}   \\
& &Ent &0.2207 & 0.0945   \\
\cline{2-5}
& \multirow{2}{*}{Mutate}& Major & 0.1413 & \textbf{0.0256}   \\
& &Ent &\textbf{0.1348} & 0.0446 \\
\hline
\end{tabular}
\end{table}

Table~\ref{errorJS} shows the L1 and L2 distance between the word count curve of $\mathcal{D}_{\text{Test}}$ with $\mathcal{D}_{\text{Major}}$ and $\mathcal{D}_{\text{Ent}}$ from Figure~\ref{count of BG1}. However for dataset JS, the distance between $\mathcal{D}_{\text{Test}}$ and $\mathcal{D}_{\text{Ent}}$ is smaller than the distance with $\mathcal{D}_{\text{Ent}}$. This is because word counts does not take into account of the semantic information in the articles.

\begin{table}[ht]
\centering
  \caption{Word Count Distance for ($\mathcal{D}_{\text{Test}}$, $\mathcal{D}_{\text{Major}}$) and ($\mathcal{D}_{\text{Test}}$, $\mathcal{D}_{\text{Ent}}$)}
  \label{errorJS}
  
\begin{tabular}{c|c|c|c}
 \hline
Dataset  &  & L1 norm & L2 norm \\
 \hline
JS   & Major  & 9.75e+3&2.90e+8 \\
   &error($\mathcal{D}_{\text{Ent}}$, $\mathcal{D}_{\text{Test}}$)  & \textbf{9.67e+3}&\textbf{2.37e+8}  \\
 \hline
IP   & Major, $\mathcal{D}_{\text{Test}}$) &\textbf{4.30e+3}&\textbf{3.90e+7}  \\
   &Ent, $\mathcal{D}_{\text{Test}}$) &4.62e+3 &4.71e+7	 \\
 \hline
\end{tabular}
\end{table}

\subsection{Analyzing Initial Triples}

Clearly, the choice of initial triples, $\mathcal{I}$ is critical in obtaining meaningful induced multivariate count time series. Instead of restricting $\mathcal{I}$ to a small number of initial triples from selected news outlets, in this section, we will consider the \emph{entire} set of initial triples collected from articles published on the first and second day of 27 randomly selected news outlets. A news initial set of triples $\mathcal{I}$ was collected from the articles published on the first and second day. We took articles from 27 randomly selected news outlets. For the dataset JS, $|\mathcal{I}| = 1216$. For the dataset IP, $|\mathcal{I}| = 570$. 

In these large sets of initial triples, we observed that the induced time series have very different dynamics. We applied K-means clustering algorithm with Euclidean distance to $\mathbf{y}^{(i)}(n), i \in \mathcal{I}$ \cite{aghabozorgi2015time}. This decomposes the set of initial triples into $K$ clusters:
\[
\mathcal{I} = \mathcal{I}_1 \cup \mathcal{I}_2 \cup \ldots \cup \mathcal{I}_K.
\]
The discrete-time Hawkes process is then learned based on the time-series, $\mathbf{y}^{(i)}(n)$, induced by each cluster of initial triples. By analyzing the distribution of daily arrivals of the different change types, we decided on $K=3$. For dataset JS, the number of initial triples in each cluster are $|\mathcal{I}_1| = 155, |\mathcal{I}_2| = 851, |\mathcal{I}_3| = 210$. For dataset IP, the number of initial triples in each cluster are $|\mathcal{I}_1| = 127, |\mathcal{I}_2| = 388, |\mathcal{I}_3| = 55$.

\begin{figure*}[htb]
     \centering
     \begin{subfigure}{0.45\textwidth}
         \centering
         \includegraphics[width=1.0\textwidth]{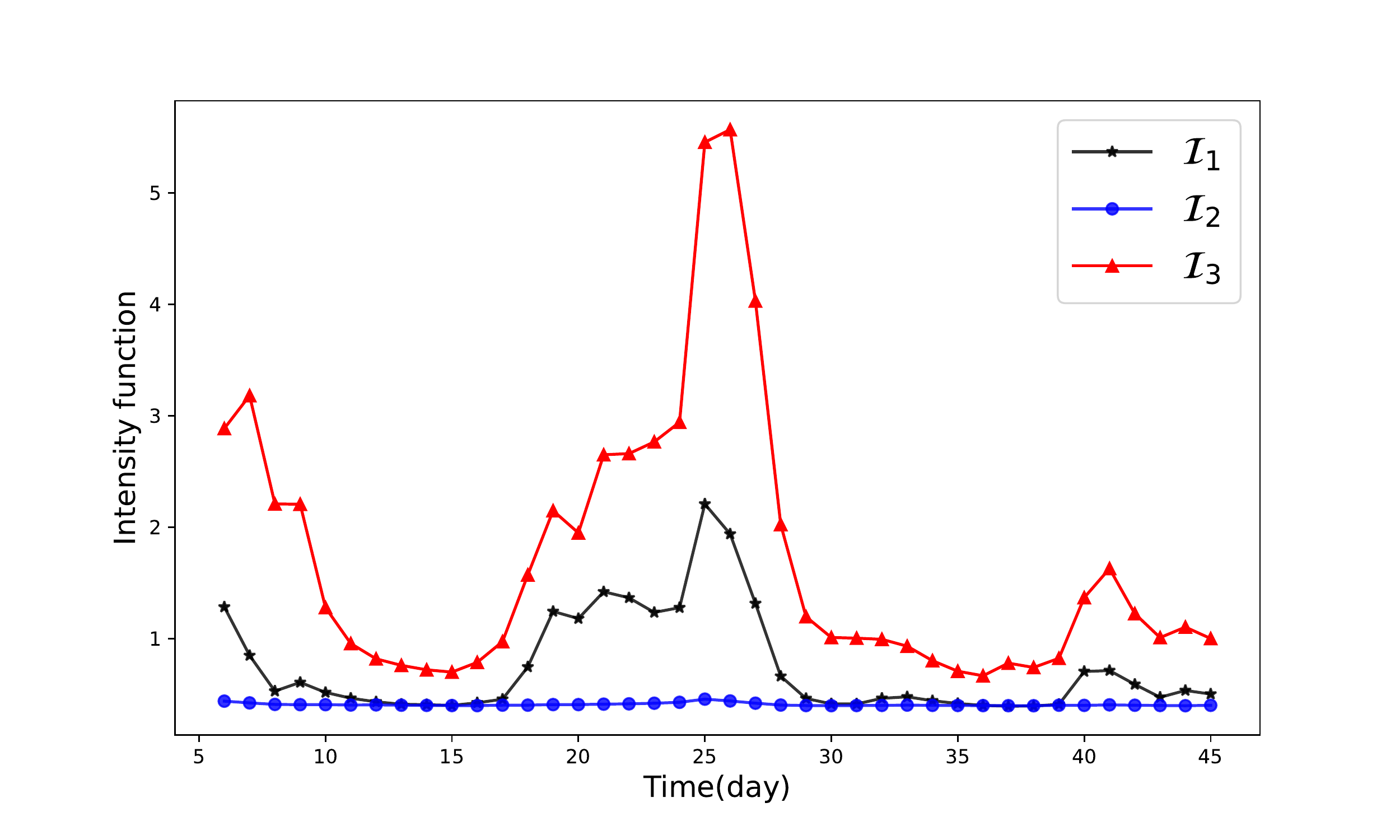}
         \caption{$\overline{\lambda_{\text{append}}}(n)$ of dataset JS}
         \label{JS lambda c1}
     \end{subfigure}
     \begin{subfigure}{0.45\textwidth}
         \centering
         \includegraphics[width=1.0\textwidth]{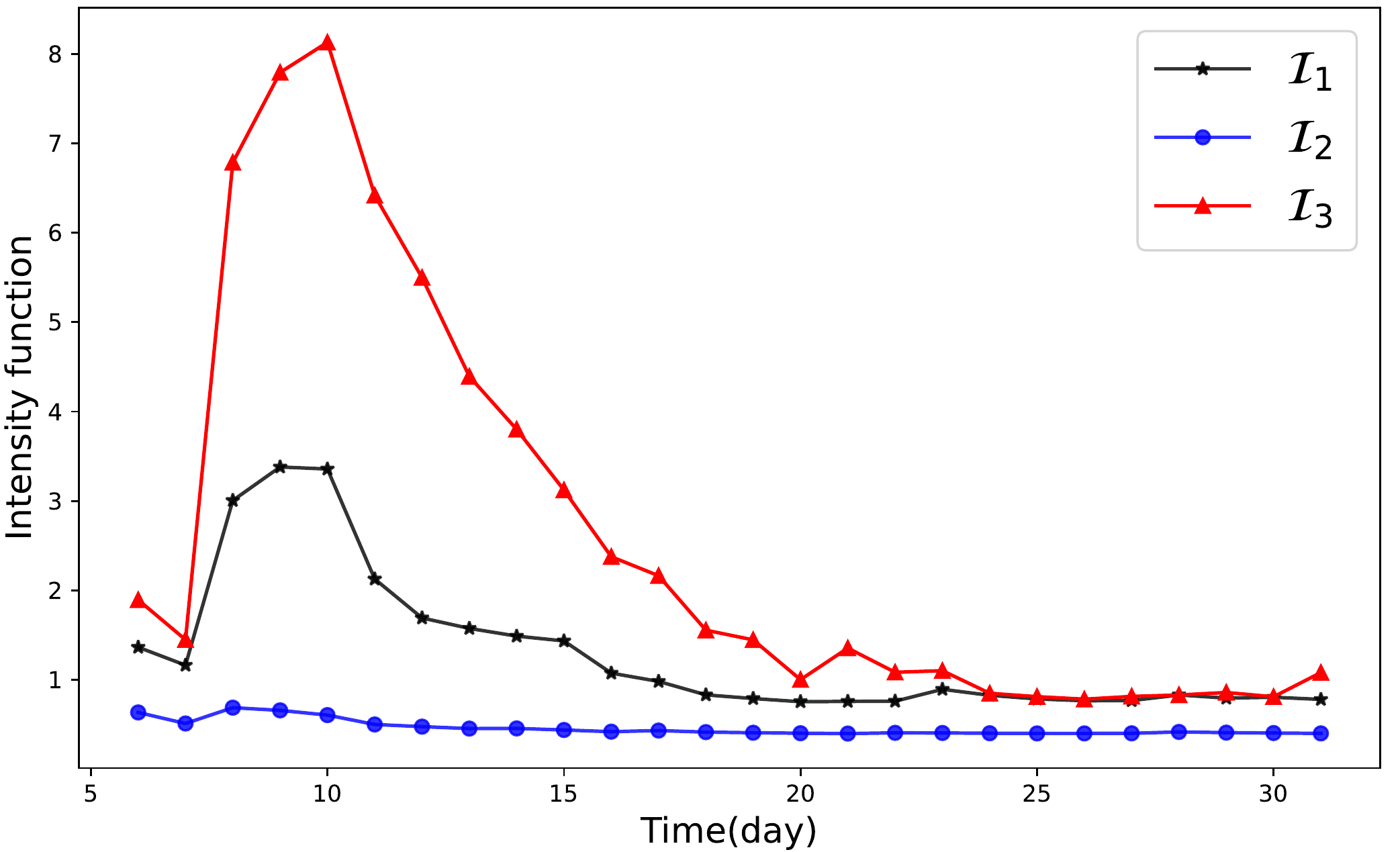}
         \caption{$\overline{\lambda_{\text{append}}}(n)$ of dataset IP}
         \label{JS lambda c2}
     \end{subfigure}
    \hfill
     \begin{subfigure}{0.45\textwidth}
         \centering
         \includegraphics[width=1.0\textwidth]{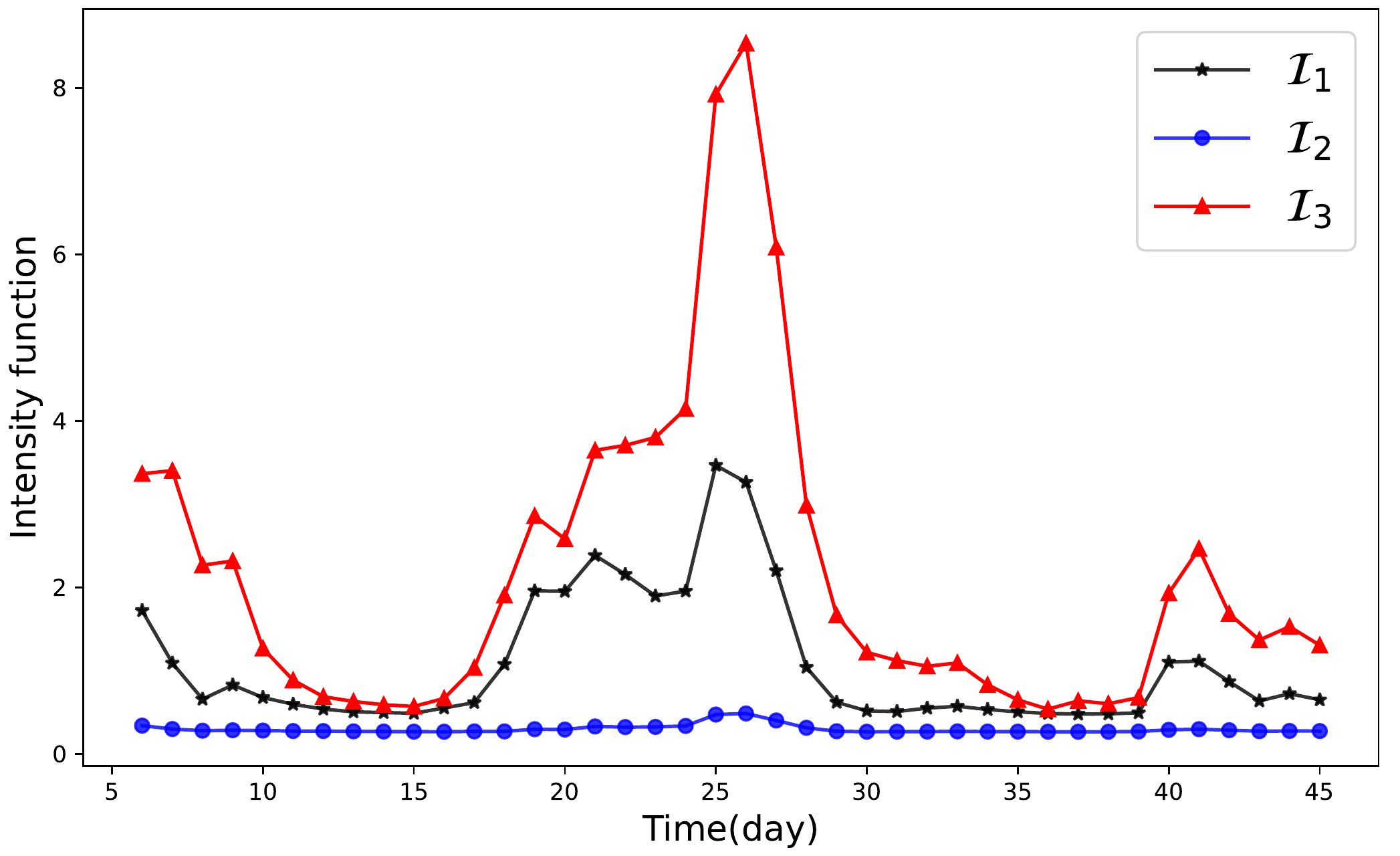}
         \caption{$\overline{\lambda_{\text{extend}}}(n)$ of dataset JS}
         \label{JS lambda c3}
     \end{subfigure}
     \begin{subfigure}{0.45\textwidth}
         \centering
         \includegraphics[width=1.0\textwidth]{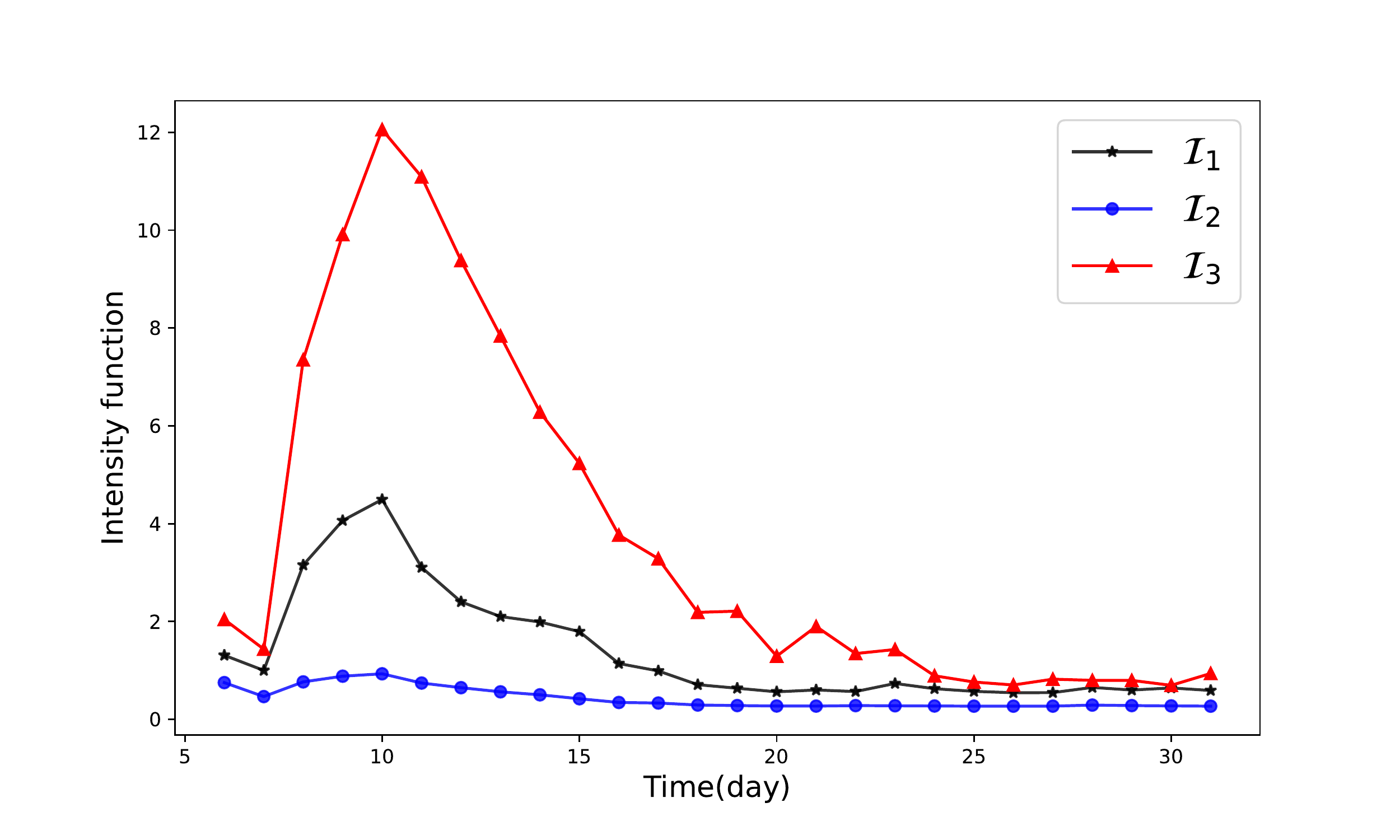}
         \caption{$\overline{\lambda_{\text{extend}}}(n)$  of dataset IP}
         \label{IP lambda c1}
     \end{subfigure}
    \hfill
     \begin{subfigure}{0.45\textwidth}
         \centering
         \includegraphics[width=1.0\textwidth]{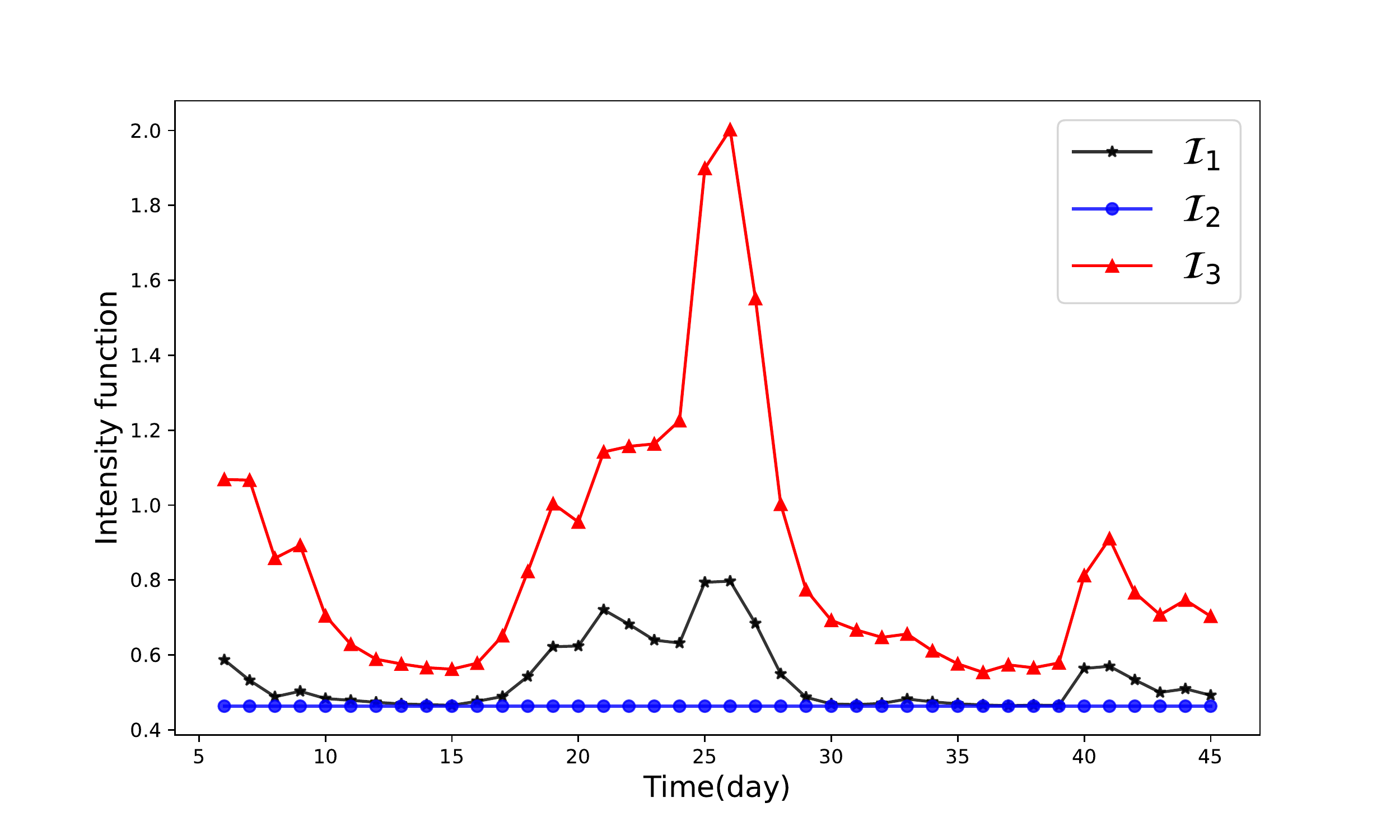}
         \caption{$\overline{\lambda_{\text{mutate}}}(n)$ of dataset JS}
         \label{IP lambda c2}
     \end{subfigure}
     \begin{subfigure}{0.45\textwidth}
         \centering
         \includegraphics[width=1.0\textwidth]{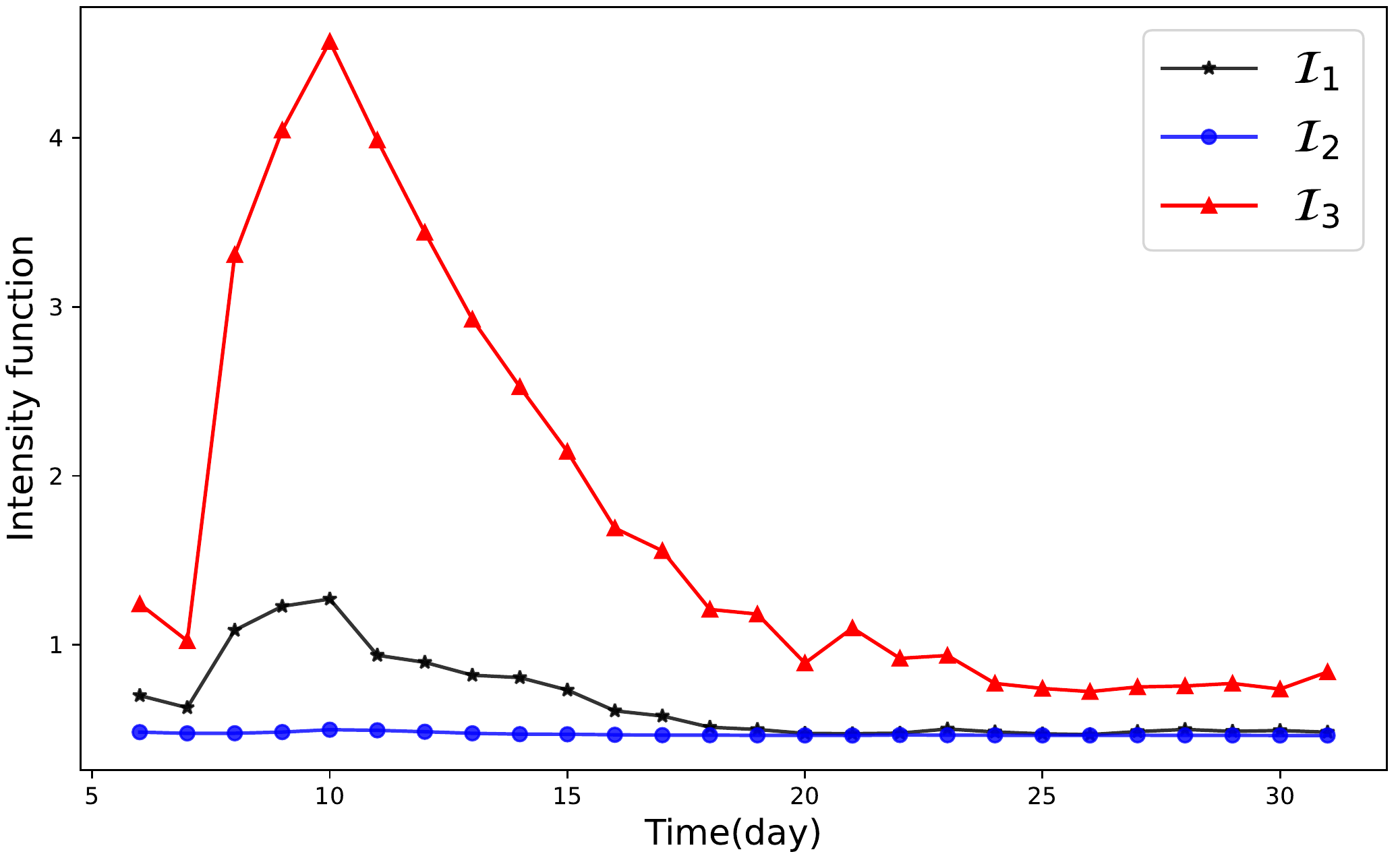}
         \caption{$\overline{\lambda_{\text{mutate}}}(n)$ of dataset IP}
         \label{IP lambda c3}
     \end{subfigure}     
        \caption{Average Conditional Intensity Function $\overline{\lambda^{(\mathcal{I}_k)}}(n)$ }
        \label{lambda graphs}
\end{figure*}

Figure~\ref{lambda graphs} shows the average intensity function, $\overline{\lambda_m^{(\mathcal{I}_k)}}(n)$ for append, extend, and mutate events for the three clusters. We note that the conditional expected number of append, extend, and mutate is essentially zero for $\mathbf{y}^{(i)}(n), i \in \mathcal{I}_2$. This means that $\mathcal{I}_2$ contains semantic triples that are not discussed over time.

\subsection{RDF Graphs of Initial Triples}

\begin{figure*}[!ht]
     \centering
     \begin{subfigure}{0.6\columnwidth}
         \centering
         \includegraphics[width=1.0\textwidth]{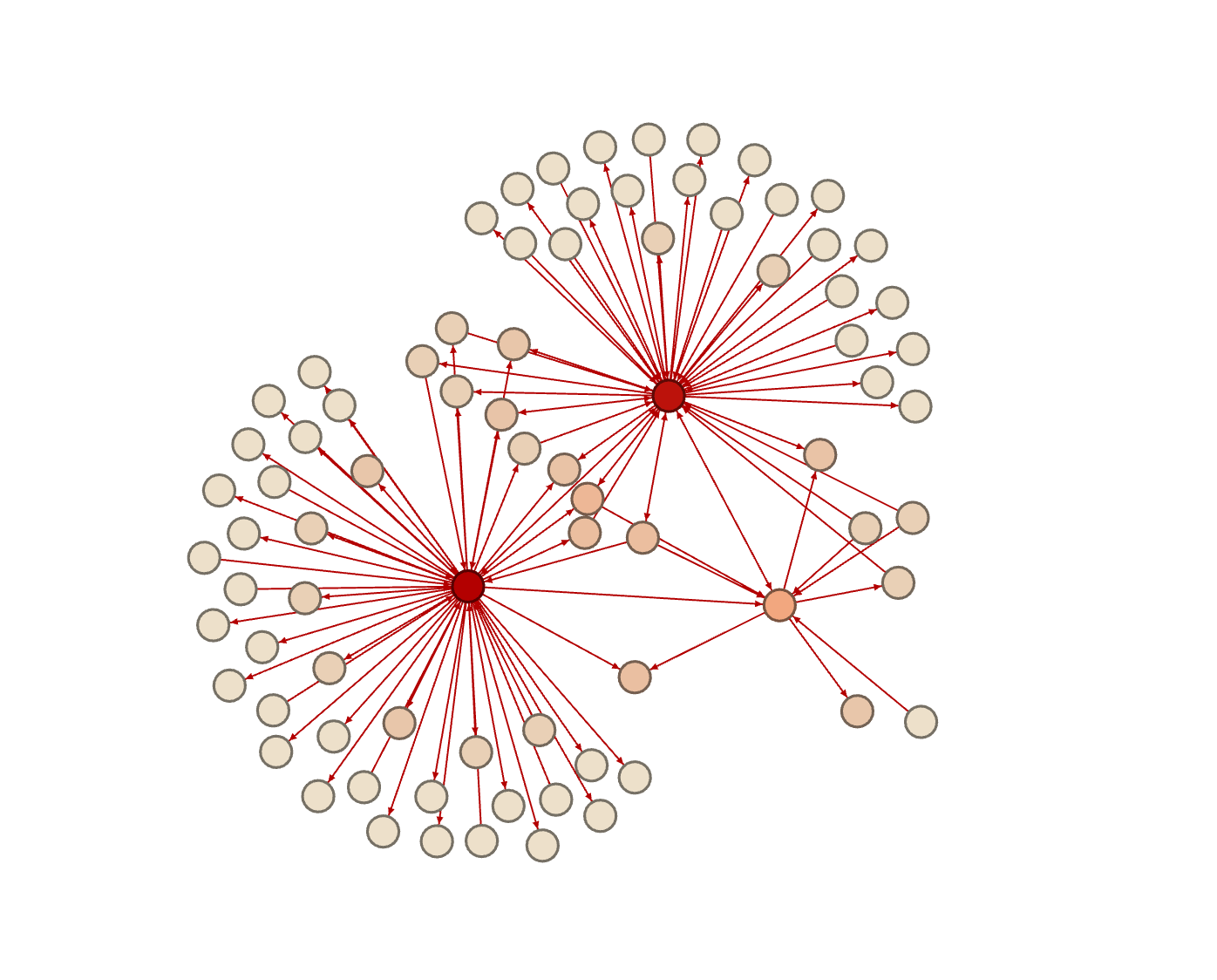}
         \caption{$G^{(\mathcal{I}_1)}(1)$ of dataset JS, $|\mathcal{I}_1| = 155$}
         \label{JS subgraph c1}
     \end{subfigure}
     \hfill
     \begin{subfigure}{0.6\columnwidth}
         \centering
         \includegraphics[width=1.0\textwidth]{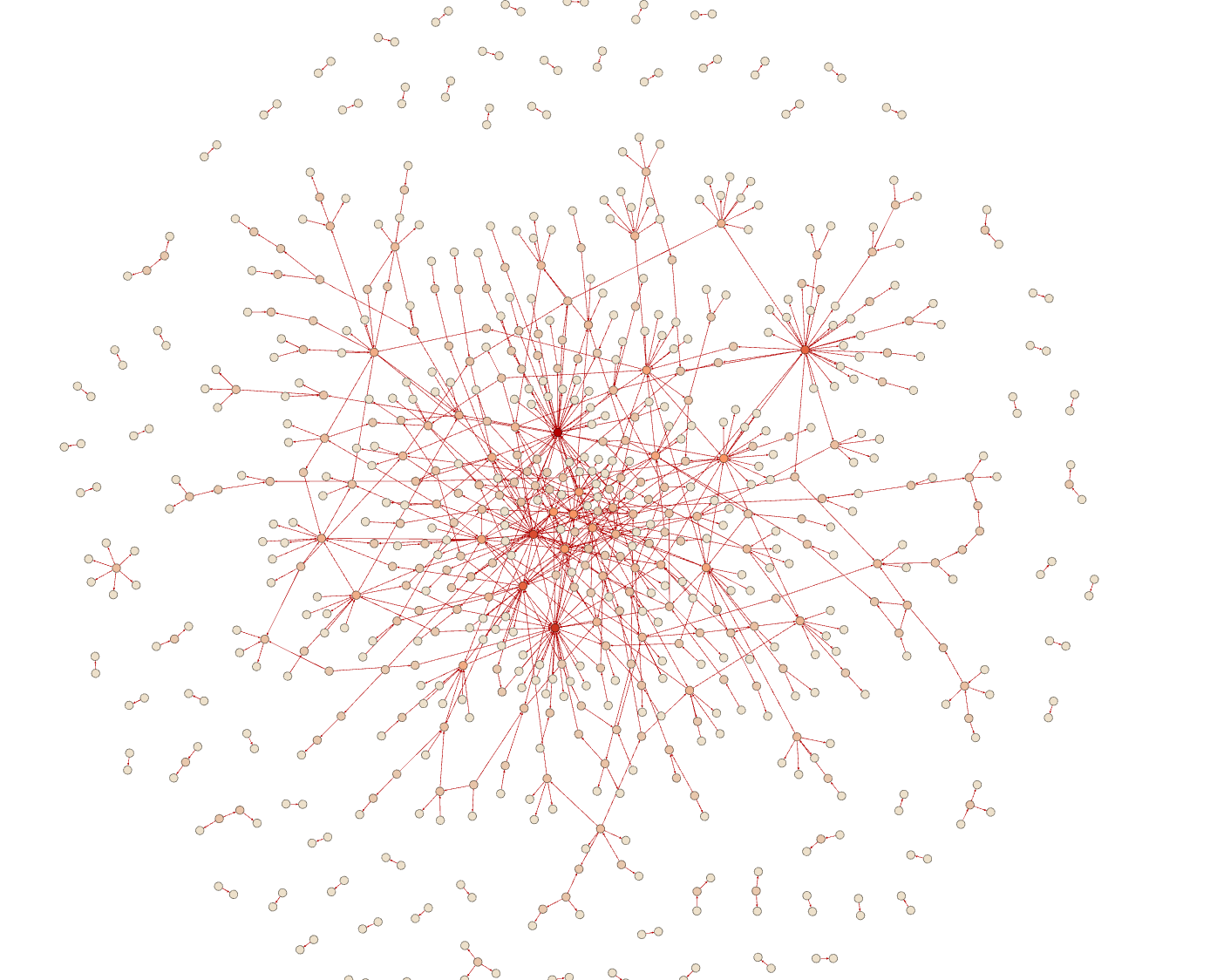}
         \caption{$G^{(\mathcal{I}_2)}(1)$ of dataset JS, $|\mathcal{I}_2| = 851$}
         \label{JS subgraph c2}
     \end{subfigure}
     \hfill
     \begin{subfigure}{0.6\columnwidth}
         \centering
         \includegraphics[width=1.0\textwidth]{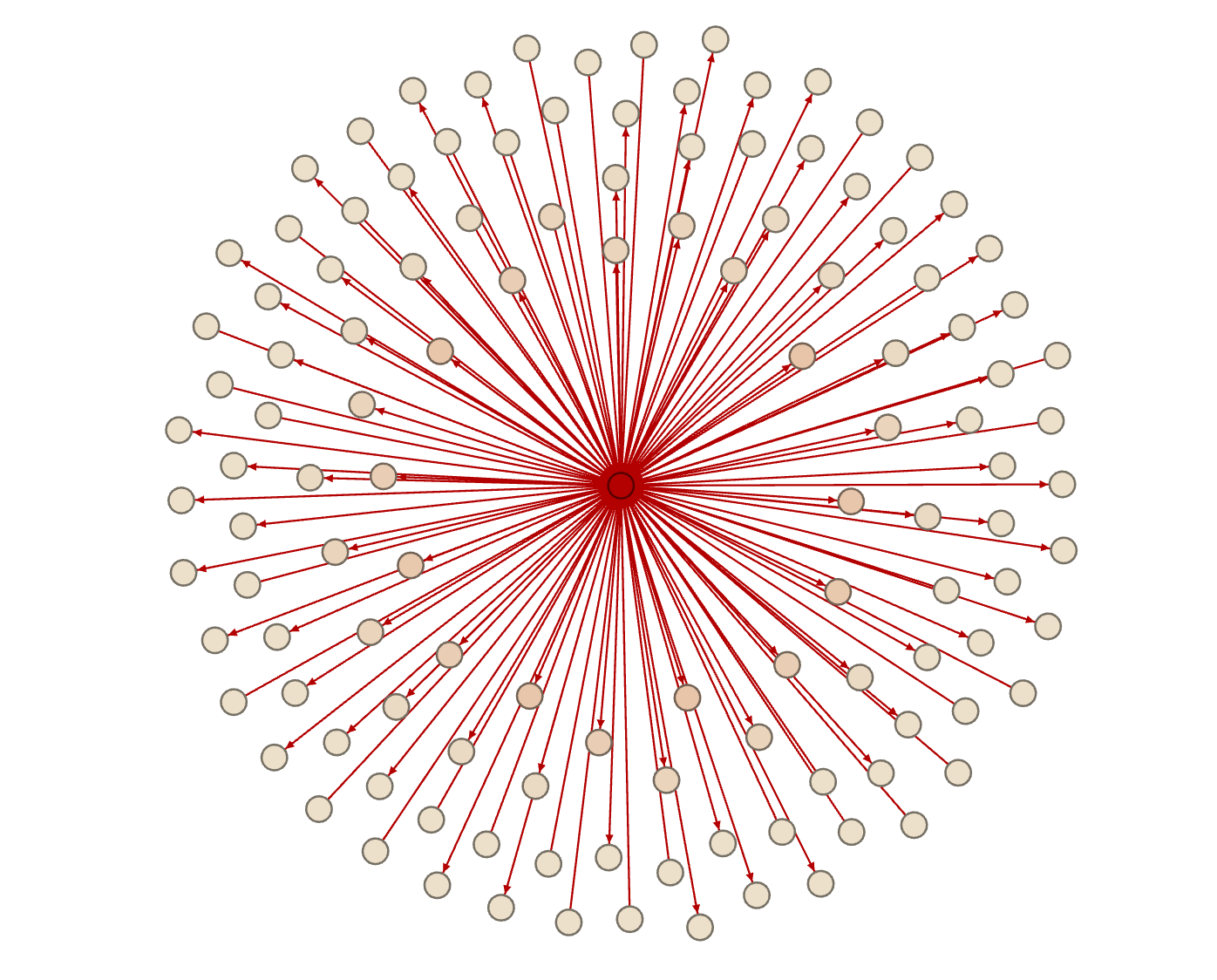}
         \caption{$G^{(\mathcal{I}_3)}(1)$ of dataset JS, , $|\mathcal{I}_3| = 210$}
         \label{JS subgraph c3}
     \end{subfigure}
     \hfill
     \begin{subfigure}{0.6\columnwidth}
         \centering
         \includegraphics[width=1.0\textwidth]{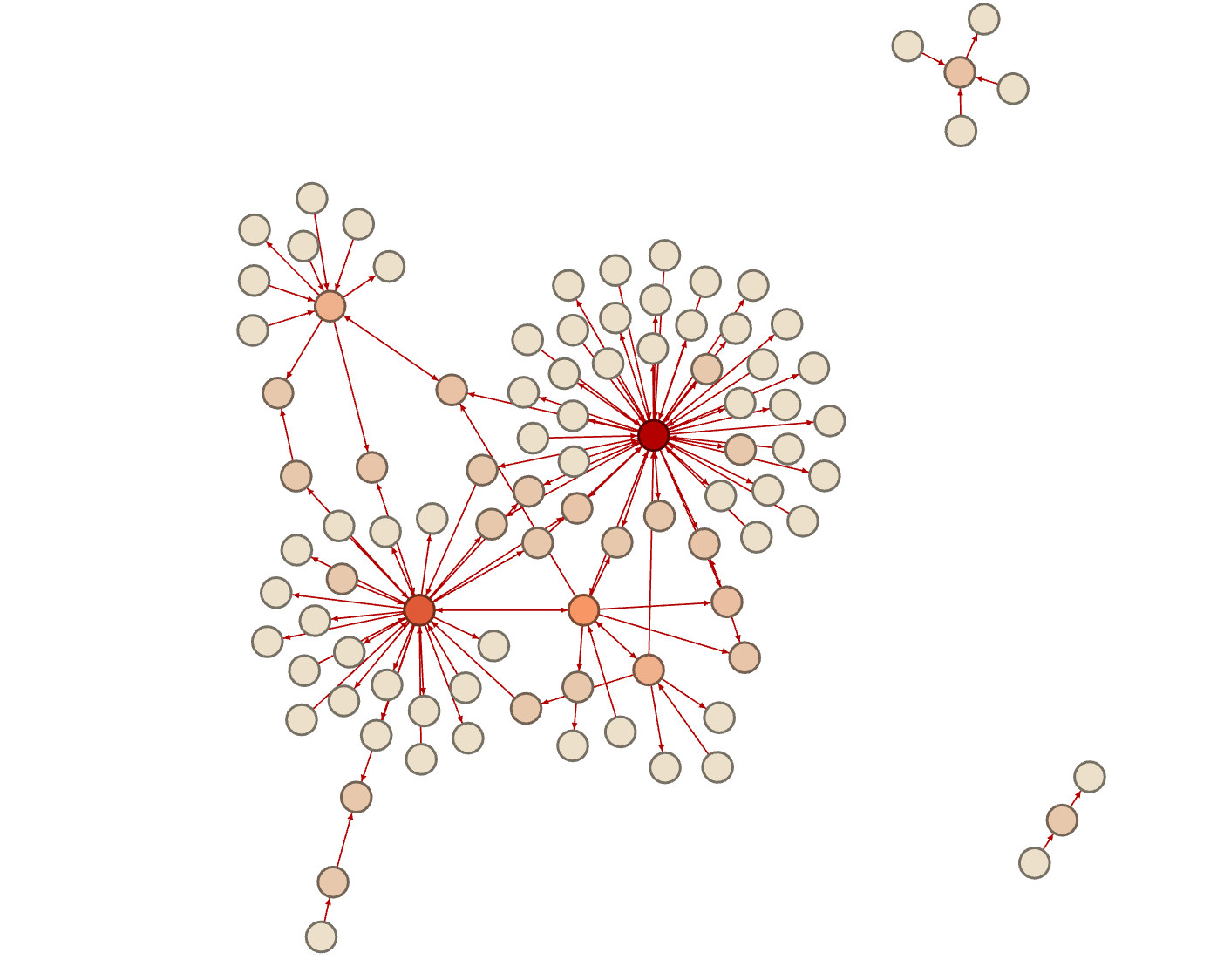}
         \caption{$G^{(\mathcal{I}_1)}(1)$ of dataset IP, $|\mathcal{I}_1| = 127$}
         \label{IP subgraph c1}
     \end{subfigure}
     \hfill
     \begin{subfigure}{0.6\columnwidth}
         \centering
         \includegraphics[width=1.0\textwidth]{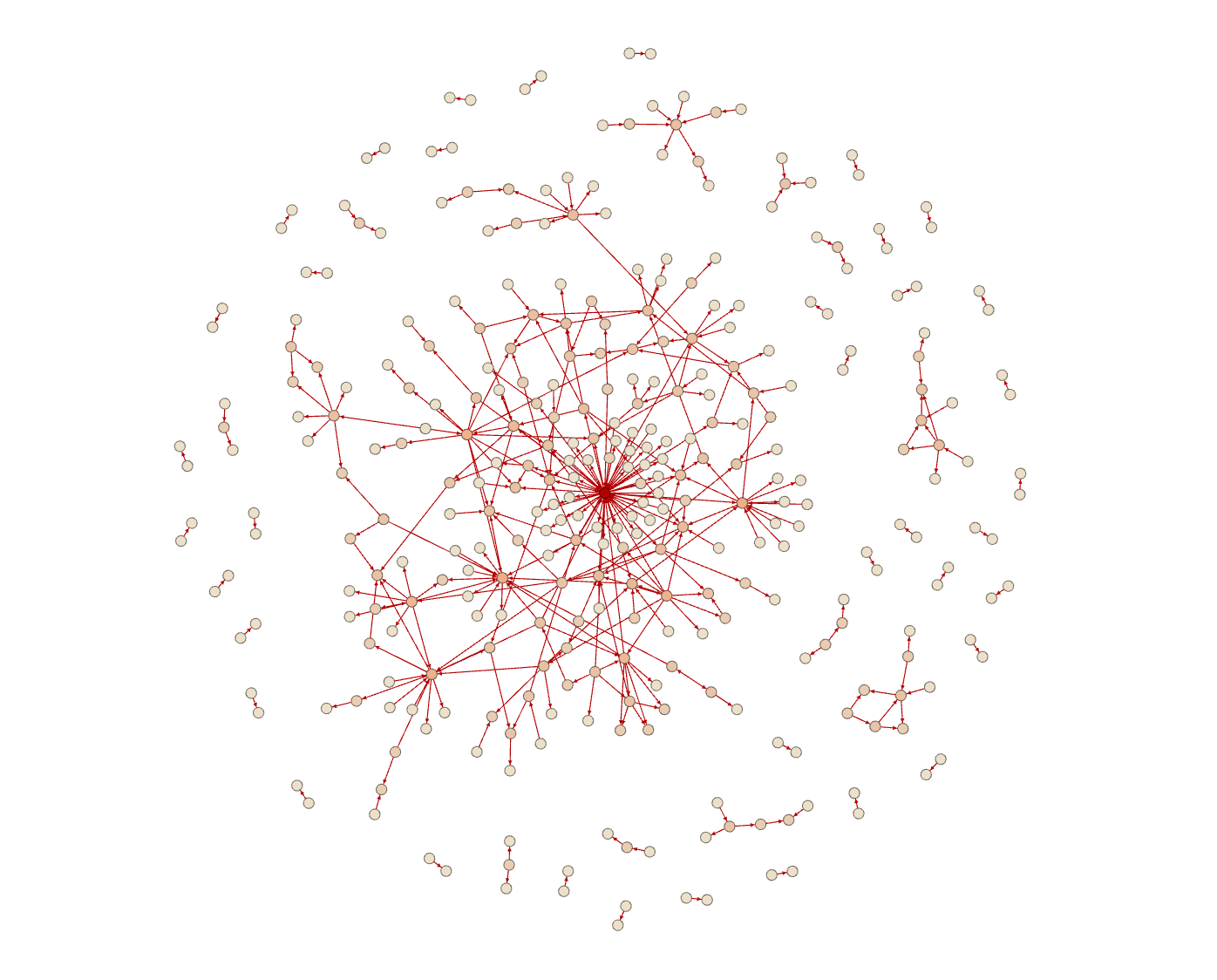}
         \caption{$G^{(\mathcal{I}_2)}(1)$ of dataset IP, $|\mathcal{I}_2| = 388$}
         \label{IP subgraph c2}
     \end{subfigure}
     \hfill
     \begin{subfigure}{0.6\columnwidth}
         \centering
         \includegraphics[width=1.0\textwidth]{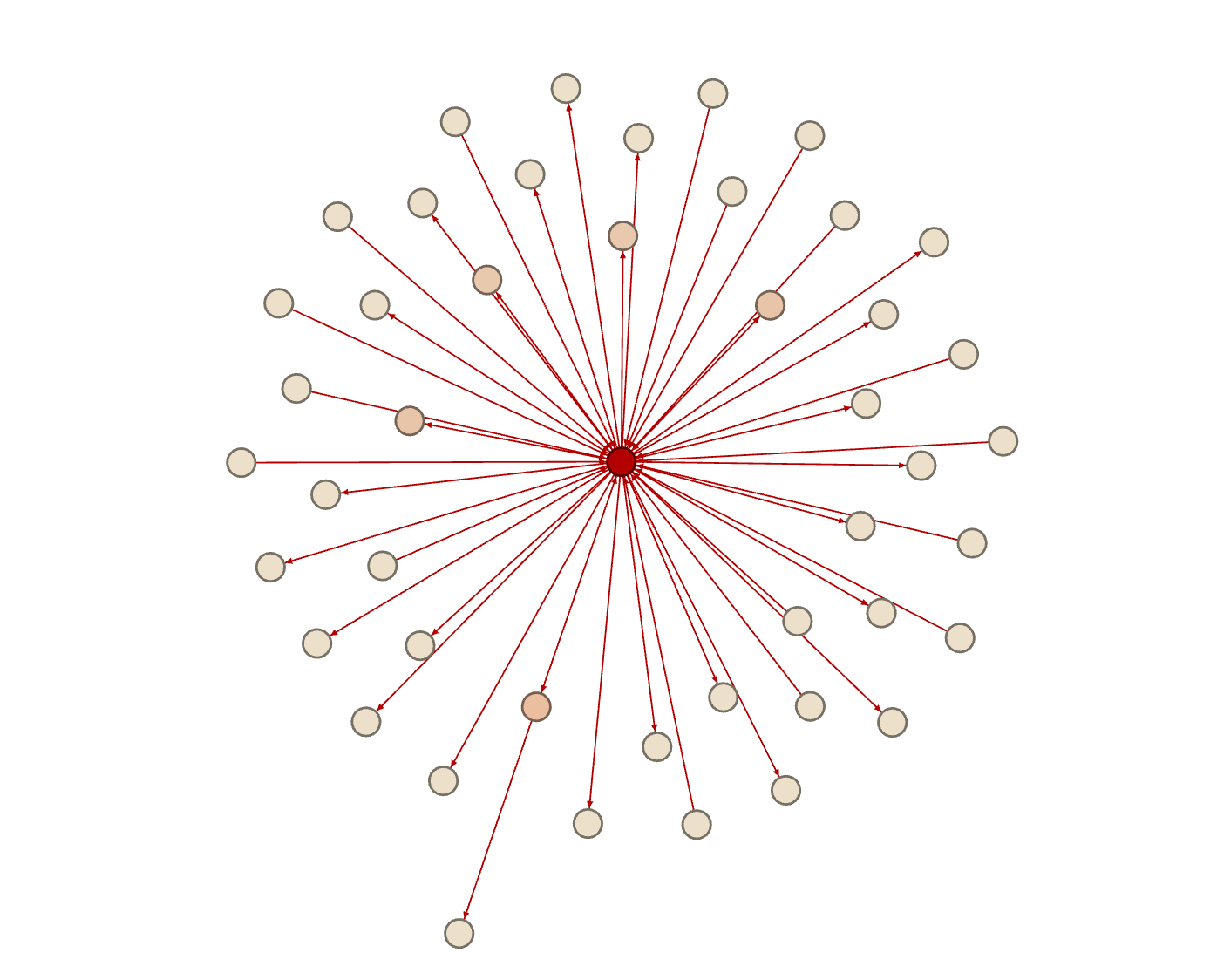}
         \caption{$G^{(\mathcal{I}_3)}(1)$ of dataset IP, $|\mathcal{I}_3| = 55$}
         \label{IP subgraph c3}
     \end{subfigure}
        \caption{RDF graphs constructed from different set of initial triples (words and multi-edges are removed)}
        \label{IPJS subgraphs}
\end{figure*}

To investigate why different clusters of initial triples lead to different dynamic behavior, we can visualize the RDF graph generated by the triples. Let $G^{(\mathcal{I}_1)}(1)$ denote the RDF graph generated on day one from the triples in $\mathcal{I}_1$; $G^{(\mathcal{I}_2)}(1)$ to denote the RDF graph generated on day one from the triples in $\mathcal{I}_2$, etc. Figure~\ref{IPJS subgraphs} shows three RDF graphs. We can also see that the RDF graphs are topologically different for $\mathcal{I}_1, \mathcal{I}_2, \mathcal{I}_3$.

The RDF graphs associated with $G^{(\mathcal{I}_1)}(1)$ have a multi-hub structure. The hubs consist of important subjects related to the news event such as \textit{police, two man, actor} for dataset JS and \textit{donald trump, united states, soleimani} for dataset IP. On the other hand, $G^{(\mathcal{I}_2)}(1)$ is more decentralized and contains many isolated triples. As shown in Table~\ref{table keywords1}, these isolated triples tend to reflect phrases that are tangentially (e.g., \textit{jamal lyon, maga country, lee daniels, trump} for dataset JS and \textit{state department, secretary mark esper} for dataset IP) related to the main subject. Since many of these triples are only tangentially related to the events being covered, they do not grow (via append, extend, mutate) over time; $G^{(\mathcal{I}_2)}(1)$ remains disconnected over time. We can see that $G^{(\mathcal{I}_3)}(1)$ are star networks. For the dataset JS, the central node is the phrase \textit{jussie smollett}. For the dataset IP, the central node is the phrase \textit{iran}. 

\begin{table}[hbt!]
\centering
  \caption{Words Associated to the nodes of $G^{(\mathcal{I}_k)}(1)$ in dataset JS}
\begin{tabular}{|p{2.5cm}|p{2.5cm}|p{2.5cm}|}
 \hline
$\mathcal{I}_1$ & $\mathcal{I}_2$ & $\mathcal{I}_3$\\
 \hline
 \hline
police, two men, actor, attack, empire, surveillance video, hate crime, unknown chemical substance, racial homophobic slurs
 
 &actor, attack, empire, people, fox, jamal lyon, maga country, unknown chemical substance, racial homophobic slurs, lee daniels,trump 
 
 &jussie smollett, two men, empire, northwestern memorial hospital, chicago,attackers \\
 \hline
\end{tabular}
\label{table keywords1}
\end{table}

\begin{table}[hbt!]
\centering
  \caption{Words Associated to the nodes of $G^{(\mathcal{I}_k)}(1)$ in dataset IP}
\begin{tabular}{|p{2.5cm}|p{2.5cm}|p{2.5cm}|}
 \hline
$\mathcal{I}_1$ & $\mathcal{I}_2$ & $\mathcal{I}_3$\\
 \hline
 \hline
donald trump,
united states,
soleimani,
people,
officials, 
tehran,
plane,
strike
 
 &soleimani, 
 attack, 
 american troops, 
 state department, 
 iranian forces, 
 iran s supreme national security council,
 defence secretary mark esper, 
 iran s top general
 
 &iran, 
 soleimani, 
 sanctions, 
 retaliation,
 baghdad, 
 mahmoud ahmadinejad \\
 \hline
\end{tabular}
\label{table keywords2}
\end{table}

\section{Conclusion}\label{sec:con}
We proposed a framework of converting a corpus of news articles collected over time to a sequence of RDF graphs using semantic triples extraction. It is very challenging to model dynamic graphs. Therefore, we considered an approximation by using multivariate count time series, which kept track of three types of topological changes to the set of initial triples in the sequence of RDF graphs: append, extend, mutate. We then fit the time series using the discrete-time Hawkes process. Analyzing the multivariate count time series gives us insights into both the dynamics and semantics of the news coverage.

The choice of the set of initial triples is extremely important. Furthermore, triples were often too specific to the exact phrasing of the sentence. For future work, we will investigate using more conceptual phrases rather than semantic triples to induce the set of time series. Our current method do not incorporate the relationships amongst the initial triples so we can make the independence assumption in our modeling. For future work, we will consider models that can account for the dependence amongst the initial triples.

\section*{Acknowledgment}
This work was funded in part by the Defense Advanced Research Projects Agency (DARPA) Active Interpretation of Disparate Alternatives (AIDA) Program under Air Force Research Laboratory (AFRL) prime contract no. FA8750-19-2-0027. Any opinions, findings, and conclusion or recommendations expressed in this material are those of the authors and do not necessarily reflect the view of the DARPA, AFRL, or the US government. This work was also partially supported by the National Science Foundation AI Institute in Dynamic Systems (Grant No. 2112085)

\bibliographystyle{ACM-Reference-Format}
\bibliography{main_ref}
\clearpage

\appendix

\section{APPENDIX}
\subsection{Data clean detail}\label{A1:data detail}

The code of how we clean the duplicate phrase has been uploaded into Github \url{https://github.com/honggen-zhang/News-Evolve-on-DHP}. Table \ref{coarse} and Table \ref{fine} list part of results of duplicate phrases defined by similarity functions in section\ref{subsec: data clean}.
\begin{table}[hbt!]
\centering
  \caption{Duplicate Phrases defined by Coarse Similarity Function}
  \label{coarse}
\begin{tabular}{p{1.5cm}|p{5.5cm}}
 \hline
key & duplicate phrases\\
 \hline
2020 election & 2020 presidential election, 2020 general election\\
 \hline
african american family & african american, african american culture, african american causes, gay african american, african american gay, african american students, african american man, african american go\\
 \hline
cardi b &alamy cardi b
3 cardi b,
getty cardi b,
friend cardi b,
cardi bcardi b\\
 \hline
cast members & several cast members,
number cast members,
angry cast members,
empire cast members\\
 \hline
chicago pd &
chicago pd spokesperson,
chicago pd sources,
party chicago pd,
chicago pd tmz,
chicago pd investigation,
chicago pd detectives,
chicago pd superintendent,
chicago pd repeat\\
\hline
be face & face be,
be now face,
be currently face,
may be face,
be also face\\
\hline
be fill with & fill with,
be fill up,
be fill to,
be partially fill with\\
\hline
...&...\\
\hline
\end{tabular}
\end{table}

\begin{table}[hbt!]
\centering
  \caption{Duplicate Phrases defined by Fine Similarity Function}
  \label{fine}
\begin{tabular}{p{1.5cm}|p{5.5cm}}
 \hline
key & duplicate phrases\\
 \hline
abel & brother abel,
nigerian brothers abimbola abel\\
 \hline
agree to train & participate as,
have access to,
be not require in,
participate\\
 \hline
attempt to gain b &gain,manipulate,have gain,gain in\\
 \hline
discrimina-
tion & form discrimination,
frequent violence discrimination\\
 \hline
jussie smollett &
empire actor\\
\hline
be send to & receive,
send,
have receive,
be mail to,
have send,
send out,
be receive at\\
\hline
...&...\\
\hline
\end{tabular}
\end{table}

\subsection{Common Triples}\label{A2: common triple}
Table \ref{common striple} shows the triples of generating our count time series data. We can extract them from both the major dataset and the entertainment dataset.

\begin{table}[hbt!]
\centering
  \caption{The common triples in Major news and Entertainment}
  \label{common striple}
\begin{tabular}{p{0.5cm}|p{2cm}|p{2.5cm}|p{2cm}}
 \hline
{}&Head & Relation  & Tail\\
 \hline
\multirow{8}{*}{JS}& chicago police&say&statement\\
\cline{2-4}
 &congress&have a lot of&urgency\\
\cline{2-4}
&detectives&be currently work to gather&video\\
\cline{2-4}
&jussie smollett&be in&chicago\\
\cline{2-4}
&jussie smollett&play&jamal lyon\\
\cline{2-4}
&tmz&first report&news\\
\cline{2-4}
&two men&pour&racial homophobic\\
\cline{2-4}
&...&...&...\\
\hline
\hline
\multirow{8}{*}{IP}& iran&have&long proud history
\\
\cline{2-4}
 &iranian people&endure&pro western regime shah\\
\cline{2-4}
&iranians&kill&united states spy plane\\
\cline{2-4}
&soleimani&have just disembark from&plane\\
\cline{2-4}
&iran&also bore&responsibility\\
\cline{2-4}
&united states&be clearly motivate by&latter s desire\\
\cline{2-4}
&truth&deny&cia\\
\cline{2-4}
&...&...&...\\
\hline
\end{tabular}
\end{table}

\subsection{News Outlets}\label{A3: news outlets}
Table \ref{news outlest} shows the major outlets, entertainment outlets, and the outlets for test data.\\

\
\\
\
\\

\begin{table*}[hbt!]
\centering
  \caption{News outlets}
  \label{news outlest}
\begin{tabular}{c|c|c|c|c}
 \hline
Major Outlets of JS&Major Outlets IP &Ent Outlets of JS &Ent Outlets of IP & Test Outlets\\
 \hline
abc13& abc13&billboard&balleralert&breitbart\\
bbc& bbc&	bossip&	celebrity&nbcchicago\\
Boston&	dfw&celebrity&deadline&riverfronttimes\\
bostonherald&huffpost&deadline&	ew&abcnews\\
chicago&nbclosangeles& eonline&	globalnews&nbcdfw\\
chicagoreader&nbcmiami&	etcanada&hellomagazine&	usatoday\\
chicagotribune&npr&	etonline&hollywoodreporter&	news10\\
gothamist&nypost& ew&nymag&	observer\\
huffpost&philadelphia&extratv&people&	cnn\\
kron4&reuters&femalefirst&popculture&	newsweek\\
mercurynews&twincities&globalnews&rollingstone&	\\
nbcnews&&heroichollywood&thehollywoodunlocked&\\	
newsday&&hollywoodlife&thesun&\\
npr&&hollywoodreporter&tmz&	\\
nydailynews&&jezebel&&	\\
nypost&&	justjared&&	\\
nytimes&&	nymag&&	\\
politico&&	ohnotheydidnt&&	\\
reuters&&	pagesix&&	\\
seattletimes&&	people&&	\\
twincities&&perezhilton&&\\	
	&&popculture&&	\\
	&&popsugar&	&\\
	&&radaronline&&\\	
	&&rollingstone&&\\	
	&&socialitelife&&	\\
	&&thehollywoodgossip&&	\\
	&&thehollywoodunlocked&&	\\
	&&themarysue&&	\\
	&&theringer&&	\\
	&&thesun&&	\\
	&&tmz&&	\\
	&&toofab&&\\	
	&&usmagazine&&\\	
	&&vanityfair&&\\	
	&&variety&&	\\
	&&zimbio&&	\\
 \hline

\end{tabular}
\end{table*}

\end{document}